\documentclass{article}
\usepackage{iftex}
\ifPDFTeX
  \usepackage[utf8]{inputenc}
\fi
\ifXeTeX
  \AtBeginDocument{}
\fi
\makeatletter
\def\input@path{{latex_template/}}
\makeatother
\usepackage[compatV3]{fancyhdr}
  \usepackage[preprint]{tmlr}
  \usepackage{lmodern}
  \usepackage{charter}

\usepackage[T1]{fontenc}
\usepackage{booktabs}
\usepackage{amsfonts}
\usepackage{nicefrac}
\usepackage{microtype}
\usepackage{xcolor}
\usepackage{graphicx}
\usepackage{amsmath,amssymb,amsthm,mathtools}
\usepackage{bm}
\usepackage{enumitem}
\usepackage{xspace}
\usepackage{caption}
\usepackage{subcaption}
\usepackage{multirow}
\usepackage{wrapfig}
\usepackage{placeins}
\usepackage{hyperref}
\usepackage{url}
\usepackage[nameinlink]{cleveref}

\crefname{section}{Sec.}{Secs.}
\Crefname{section}{Sec.}{Secs.}
\crefname{subsection}{Sec.}{Secs.}
\Crefname{subsection}{Sec.}{Secs.}
\crefname{subsubsection}{Sec.}{Secs.}
\Crefname{subsubsection}{Sec.}{Secs.}
\crefname{figure}{Fig.}{Figs.}
\Crefname{figure}{Fig.}{Figs.}
\crefname{table}{Tab.}{Tabs.}
\Crefname{table}{Tab.}{Tabs.}
\crefname{equation}{Eq.}{Eqs.}
\Crefname{equation}{Eq.}{Eqs.}

\newcommand{\method}{{\fontfamily{ppl}\selectfont GenCine}\xspace}

\definecolor{bgreen}{RGB}{0,170,0}
\definecolor{bred}{RGB}{220,0,0}
\definecolor{mydarkblue}{RGB}{0,0,150}
\definecolor{Gray}{gray}{0.93}
\hypersetup{
    colorlinks=true,
    linkcolor=bred,     %
    citecolor=mydarkblue,    %
    urlcolor=magenta     %
}

  \title{Generative Cinematographer:\\Composing Camera and Object Motion in 3D}%
\author{%
  Jiahan Zhang \quad Chaohao Yang \quad Namitha Guruprasad \quad Vivekjyoti Banerjee\\[3pt]
  Trong-Tung Nguyen \quad Alan Yuille \quad Anand Bhattad\\[12pt]
  Johns Hopkins University
}

\begin{document}
\maketitle
\vspace{-0.52cm}
  \noindent\textbf{Project Website: }
  {\hypersetup{urlcolor=mydarkblue}\url{https://generative-cinematographer.github.io/}}\par%
  \begin{figure}[p]%
  \centering
      \includegraphics[width=\linewidth,height=0.82\textheight,keepaspectratio]{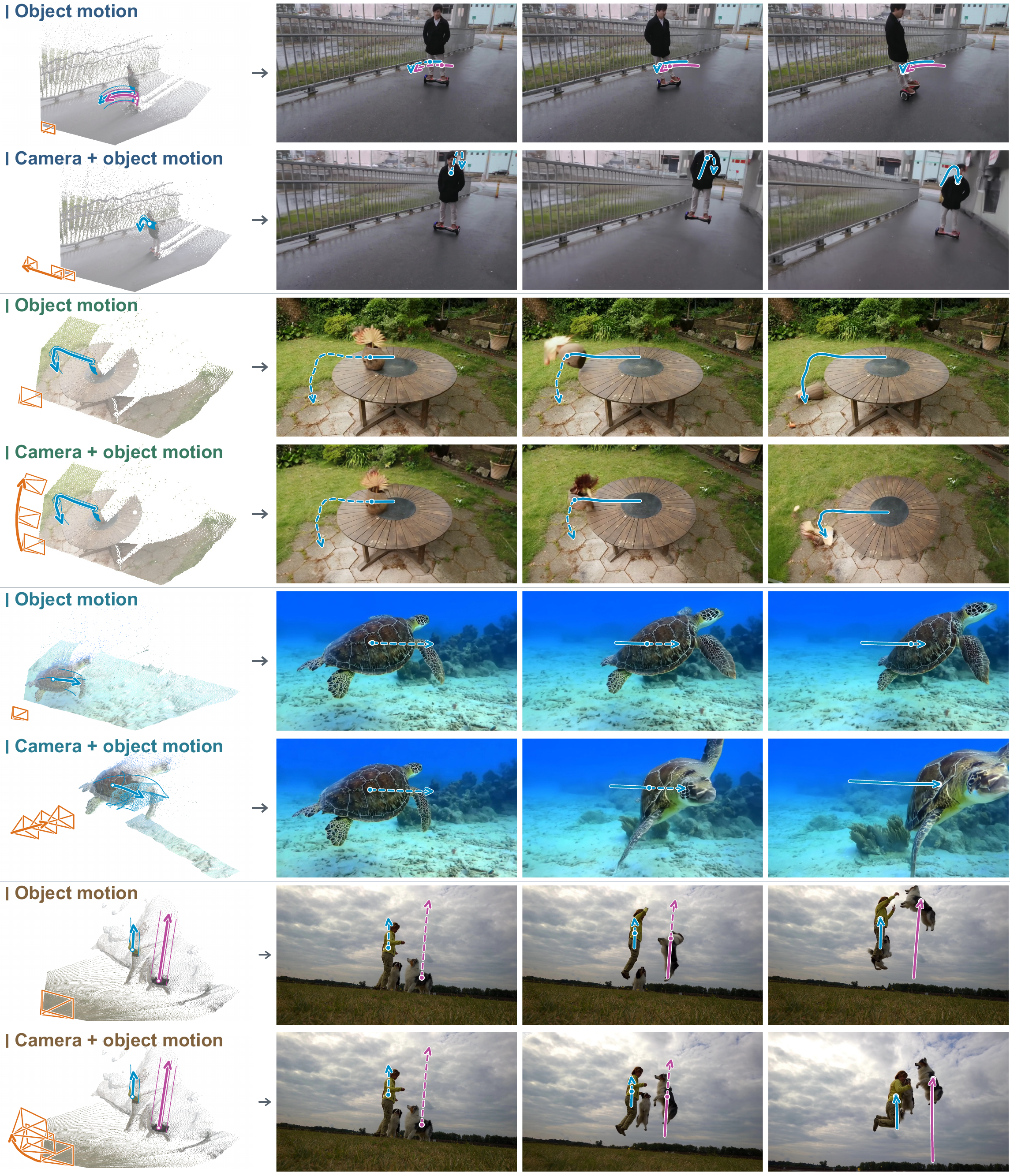}%
 \captionsetup{hypcap=false}
      \vspace{-6pt}%
\captionof{figure}{\looseness=-1 {\textbf{Composing camera and object
motion in 3D from a single image.}
Artists choreograph foreground motion and camera paths in a shared 3D
scene lifted from the input image. Each pair shows object motion with a
fixed camera (top) and combined camera and object motion (bottom).
The hoverboard rider follows a curved path or takes flight; the remaining
pairs show a potted plant moving off a table, a swimming turtle, and a
girl and dog jumping under separate controls.
Each row shows the authored 3D controls followed by three generated
frames. Orange denotes camera poses and paths; other colors denote
foreground trajectories. Solid and dashed lines indicate elapsed and
remaining motion, respectively. Best viewed as videos on the supplemental
website.}} 
\label{fig:teaser}
  \end{figure}%

  \vspace{10pt}%
\begin{abstract}
Current controllable video generation systems often rely on 2D motion
trajectories or sparse drag signals for object motion. These controls
are ambiguous because the same 2D trajectory can correspond to different
3D motions, especially when the camera and objects move simultaneously.
We present \emph{Generative Cinematographer} (\method), a system that
lifts a single image into an editable 3D scene scaffold where artists
jointly author camera and foreground motion. Artists specify a camera
path and move selected foreground regions using local 3D motion handles.
Several handles can move different parts of a subject independently,
providing a piecewise-rigid approximation to non-rigid motion without
a physics simulator or category-specific prior. To communicate these
controls to a pretrained video model, we project them into guidance maps.
These maps record where the controlled regions appear in each frame,
assign each handle a fixed color across frames and encode the
current 3D positions of its controlled points in the same world
coordinate system as the background. This lets us
describe object motion relative to the scene even as the camera moves.
For training, we recover controls from the motion observed in real
videos and use ground-truth geometry and trajectories from synthetic
videos. We train a lightweight guidance branch and LoRA adapters on a
pretrained Wan model to follow these controls. Our experiments show consistent camera-relative
motion, improved geometric consistency under viewpoint changes,
and strong controllability across diverse real-world scenes.
\end{abstract}

\section{Introduction}
\begin{figure}[t!]
  \centering
  \includegraphics[width=\textwidth,pagebox=cropbox]{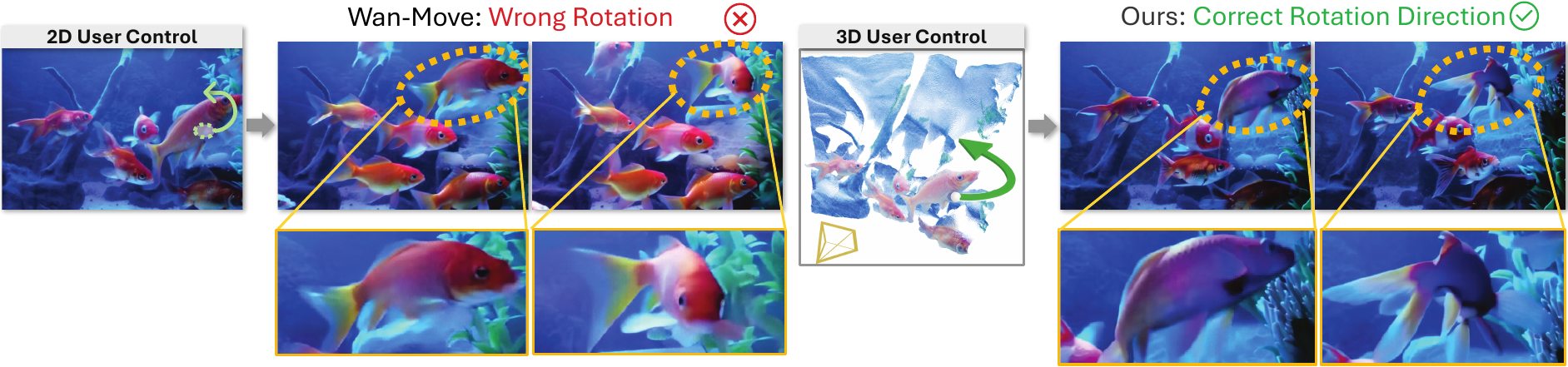}
\caption{\textbf{Why author motion in 3D?}
A 2D drag can leave the intended 3D rotation ambiguous.
Here, the 2D-controlled baseline turns the fish in a direction
inconsistent with the artist's intent, while \method follows
the rotation specified in 3D. Enlarged views highlight the
difference.}
  \label{fig:2d_w_3d}
      \vspace{-6pt}%
\end{figure}

\begin{quote}
\itshape
``Just as one can compose colors, or forms,
so one can compose motions.''
\raggedleft\\
--- Alexander Calder (1933)
\vspace{-5pt}
\end{quote}
Composing a cinematic shot means deciding how the camera and the subjects
move together. Classical graphics systems such as the Virtual
Cinematographer~\citep{he1996virtual} compose shots by controlling a camera
within a modeled 3D scene. We bring this world-space approach to video
generation, allowing artists to jointly author camera and object motion
using a 3D scaffold lifted from a single image (Fig.~\ref{fig:teaser}).
Generative video models can now synthesize plausible motion from a single
image, but giving artists precise control over these movements remains
difficult. Text prompts can describe an action but
leave its precise path unspecified. Likewise, a 2D drag may correspond
to several 3D motions or mix object movement with a change in viewpoint
(Fig.~\ref{fig:2d_w_3d}). A generated video can therefore look plausible
while following a different motion from the one the artist intended.

Many controllable video methods specify foreground motion through pixel
displacements~\citep{wu2024draganything, geng2024motionprompting}, leaving
the underlying 3D motion implicit. Physics simulators and category-specific
priors provide additional structure~\citep{liu2024physgen, li2025wonderplay},
but their applicability depends on the materials and objects they support.
Learning 3D control from videos presents a further challenge: reliable
3D motion annotations are scarce. Synthetic data provide these annotations
but differ visually from real footage, while per-pixel 2D
conditioning~\citep{chu2025wanmove} leaves artists working in image space.

A natural way to express the intended motion is to place the camera and
subjects in a shared 3D scene. Camera paths then describe how the viewpoint
changes, while object trajectories describe how the scene itself changes.
We follow this approach in \textbf{Generative Cinematographer} (\method). Starting
from a single image, we lift the scene into a colored point cloud.
The artist selects a few foreground regions and specifies how each
should move in 3D, along with a camera path. Each selected region acts
as a \emph{motion handle}: its points move together as a rigid group.
Several handles can move different parts of a subject independently,
providing a piecewise-rigid approximation to non-rigid motion without
a skeleton, physics simulator, or category-specific prior.

To train the model, we recover examples of the motion an artist might specify.
For real videos, 3D foundation models estimate geometry and camera motion.
We use these estimates to lift tracked foreground points into 3D, group
nearby tracks into local handles, and use their observed motion as
approximate 3D controls. Synthetic videos supply ground-truth geometry
and trajectories. Each clip is thus paired with controls describing the
motion that actually occurs in it; its first frame supplies the reference
image, and the clip supplies the target video.

A pretrained video model does not directly interpret a camera path or
a collection of 3D point trajectories. We need a representation that
fits its image-based input while retaining what the controls mean.
Recent methods support joint camera and object
control~\citep{zheng2026versecrafter, zhang2026symphomotion};
VerseCrafter, for example, uses RGB and depth maps of a background point
cloud and object-level Gaussian trajectories. Our representation
separates three questions: \emph{where does a region appear, which
region is it, and how does it move in the world?}
First, we view the scene through the specified camera at each frame,
placing guidance at the projected locations of background and handle
points. Second, a fixed color identifies each handle across frames.
Third, separate colors encode the current 3D positions of foreground
points and the fixed positions of background points. These form the
foreground identity, foreground XYZ, and background XYZ maps.
The two XYZ maps use the same world frame and scale: camera motion
changes where a static point appears, whereas object motion also changes
its world position. This distinction preserves information that a 2D
trajectory alone discards.

We construct these guidance sequences for each training clip and encode
them with the pretrained video model's frozen video autoencoder.
The diffusion model learns to generate the paired video from its first
frame, text, and encoded guidance. We train a lightweight side branch
and LoRA adapters while keeping the base model weights frozen, allowing
the model to learn control without retraining its visual prior.
At inference, an artist's camera path and handle trajectories pass
through the same map construction in place of recovered motion.
Varying the number and placement of training handles exposes the model
to the sparse, partial controls available during authoring.

Our contributions are:
(1) a \textbf{visual guidance representation for 3D motion} that preserves handle identity and foreground--background positions in a shared world frame while remaining compatible with pretrained video models;
(2) a \textbf{local 3D motion-handle formulation} that approximates non-rigid motion through independently controlled local rigid regions, without requiring a simulator or category-specific prior; and
(3) a \textbf{shared training and authoring pipeline} that learns from videos paired with 3D motion controls and generates videos from a single image using artist-authored camera paths and local object trajectories.

\section{Related Work}
\label{sec:related}

\paragraph{Virtual cinematography.}
\looseness=-1 The graphics community has long studied camera motion authoring. The Virtual Cinematographer~\citep{he1996virtual} formalized shot composition as cinematographic idioms. Later work extended this to declarative camera planning~\citep{christianson1996declarative} and procedural camera control~\citep{drucker1992cinema}. These systems assume a fully modeled 3D scene. In contrast, \method uses a 3D scaffold lifted from a single image to control the video generator.

\vspace{-8pt}
\paragraph{Camera and motion control in video diffusion.}
MotionCtrl~\citep{wang2024motionctrl} and CameraCtrl~\citep{he2024cameractrl, he2025cameractrl} condition generation on camera poses.
GEN3C~\citep{ren2025gen3c} renders a depth-derived 3D cache under target poses, while DimensionX~\citep{sun2025dimensionx} generates 3D and 4D scenes from one image.
WorldReel~\citep{fang2025worldreel} jointly generates RGB frames, point maps, camera trajectories, and flow for geometric and motion consistency.
Generative Camera Dolly~\citep{vanhoorick2024gcd}, TrajectoryCrafter~\citep{yu2025trajectorycrafter}, and FreeOrbit4D~\citep{cao2026freeorbit4d} redirect viewpoints in existing videos, with FreeOrbit4D reconstructing a geometry-complete 4D proxy.
Vista4D~\citep{lin2026vista4d} reshoots source videos using rendered 4D point clouds with temporally persistent static points.
MotionCtrl also supports object control; CineMaster~\citep{wang2025cinemaster} uses 3D box layouts, and Diffusion as Shader~\citep{gu2025diffusionshader} uses 3D tracking videos for joint control.
\vspace{-8pt}
\paragraph{Object motion control.}
\looseness=-1 DragNUWA~\citep{yin2023dragnuwa}, DragAnything~\citep{wu2024draganything}, and ATI~\citep{wang2025ati} control motion through image-space trajectories; other methods use entities~\citep{dragentity2024} or masks and boxes~\citep{li2025magicmotion}.
Motion-I2V~\citep{shi2024motioni2v} predicts motion before synthesis, while Wan-Move~\citep{chu2025wanmove} propagates latent features along trajectories.
Motion Prompting~\citep{geng2024motionprompting} accepts sparse or dense trajectories, and Time-to-Move~\citep{singer2025timetomove} guides a frozen generator with coarse animations.
3DTrajMaster~\citep{fu20243dtrajmaster} specifies entity-level 6DoF trajectories; MotionCanvas~\citep{xing2025motioncanvas} maps scene-aware camera and object controls to screen-space conditions.
\vspace{-8pt}
\paragraph{Joint camera and object control in 3D.}
SymphoMotion~\citep{zhang2026symphomotion} combines geometry-aware camera guidance, 2D visual signals, and 3D trajectory embeddings; VerseCrafter~\citep{zheng2026versecrafter} renders background point clouds and object Gaussian trajectories into conditioning maps.
FlexTraj~\citep{zhang2026flextraj} encodes persistent track and segmentation identities in control videos; Edit-by-Track~\citep{lee2025editbytrack} edits using paired source and target 3D tracks.
Go-with-the-Track~\citep{namekata2026gowiththetrack} uses reference-anchored tracks for compositing, camera control, and object motion.
MoRight~\citep{liu2026moright} decouples camera and object motion via canonical static-view trajectories and cross-view attention.
UniCaMo~\citep{vu2026tracknoise} constructs motion-consistent noise with occlusion-informed warping.
In \method, artist-placed local 3D handles share the camera's world frame. Mapping their persistent identities and current positions into conditioning maps jointly encodes camera and object motion.

\vspace{-5pt}
\section{Method}
\label{sec:method}
\begin{figure*}[t]
  \centering
  \includegraphics[width=\linewidth]{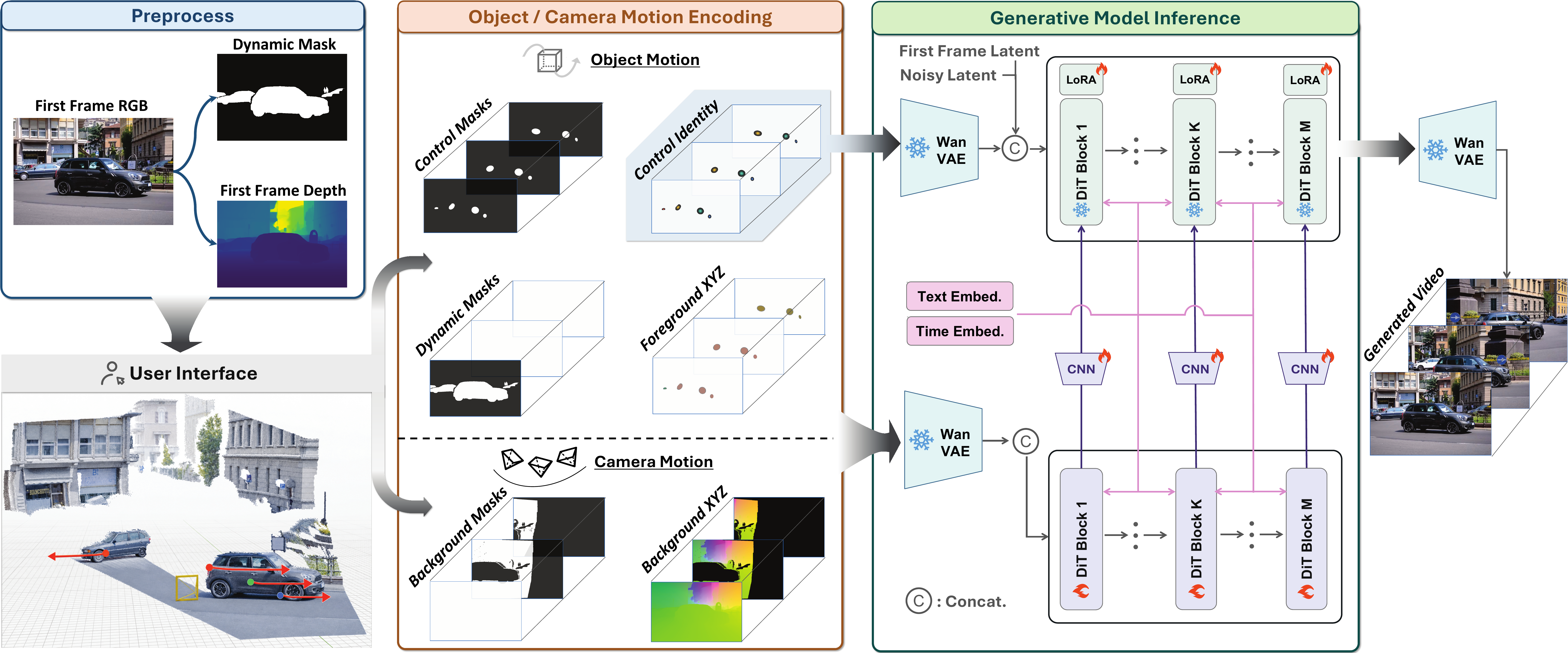}
  \caption{\textbf{The \method 3D control workflow.}
  We estimate depth and a dynamic mask from an RGB image, then lift them into a 3D scene for motion authoring.
  The artist specifies camera and object motion in a shared world frame using a 3D interface such as Blender.
  We convert the controls into background XYZ, foreground XYZ, and foreground identity maps, indicating static scene positions, moving handle positions, and handle identity.
  Validity masks distinguish available guidance from missing values, while the dynamic mask marks the initial foreground.
  A frozen Wan VAE encodes the maps for the video model. Training pairs videos with guidance recovered from their motion; at inference, the same maps express artist-authored controls.}
  \label{fig:framework_overview}
  \vspace{-15pt}
\end{figure*}

\method learns to generate a video from a reference image, a camera path,
and local foreground trajectories. Training controls are recovered from
videos; inference controls are supplied by the artist. Both pass through
the same three steps (Fig.~\ref{fig:framework_overview}):
\textbf{(i) Place the controls in each view.} Project background and handle
points through the specified camera so the model knows where guidance
applies.
\textbf{(ii) Preserve identity and 3D motion.} Use fixed handle colors to
identify regions across frames and separate XYZ maps to describe their
positions relative to the static scene.
\textbf{(iii) Learn to follow the guidance.} Encode the maps with the
frozen video autoencoder and train adaptation modules on paired
videos, reference images, and controls.

\vspace{-5pt}
\subsection{Scene Scaffold and Motion Authoring}
\label{sec:method:overview}

Given an image $\mathbf{x}_0$, we estimate depth $\mathbf{D}_0$ using
an off-the-shelf estimator~\citep{huang2025vipe} and take the initial
camera frame as the world frame. We back-project each pixel
$\mathbf{u}=(u,v)$ as
$\mathbf{p}_{\mathbf{u}}=\mathbf{D}_0(\mathbf{u})
\mathbf{K}_0^{-1}(u,v,1)^\top$,
where $\mathbf{K}_0$ is the intrinsic matrix. Attaching the image
colors yields a point cloud $\mathbf{P}$, which a predicted dynamic
mask $\mathbf{M}^{\mathrm{dyn}}$ partitions into static background
$\mathbf{P}^{\mathrm{bg}}$ and potentially mobile foreground
$\mathbf{P}^{\mathrm{fg}}$.

The artist selects foreground regions in a 3D viewport and assigns
each a trajectory. Each handle moves its selected points as a rigid
group. For $N$ handles, let $\mathcal{I}_i$ be the fixed point-index
set of handle $i$. Its world-space support at frame $t$ is
\begin{equation}
\mathcal{H}_{t,i}
=\bigl\{\mathbf{p}_{t,j}\mid j\in\mathcal{I}_i\bigr\},
\quad i=1,\ldots,N,\quad t=0,\ldots,T,
\label{eq:motion-handle-set}
\end{equation}
where $\mathbf{p}_{t,j}$ is the point's position under the authored
motion. The artist also specifies camera intrinsics and world-to-camera
poses $\{\mathbf{K}_t,\mathbf{R}_t,\mathbf{t}_t\}_{t=0}^{T}$ in the
same world frame. Supplementary Sec.~\ref{supp:scaffold} provides
scene-construction details.

\vspace{-5pt}
\subsection{Motion Representation: Correspondence Maps and Rigid Handles}
\label{sec:method:correspondence}

The motion representation in \method follows two principles: explicit correspondence and local rigidity. The first is motivated by motion correspondence in human visual perception~\citep{ullman1979interpretation,dawson1991motion}; the second allows a few local controls to approximate complex foreground deformation~\citep{sumner2007embedded,sorkine2007arap}.

\begin{figure*}[t!]
  \centering
  \includegraphics[width=\linewidth]{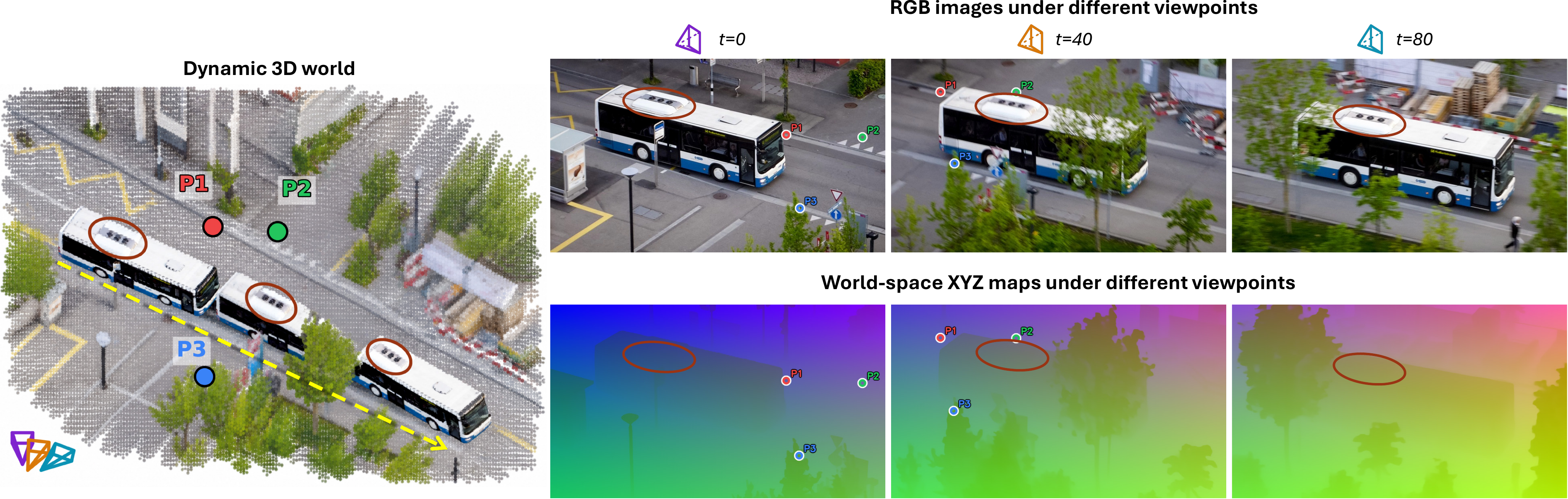}
  \caption{\textbf{World-space XYZ maps establish correspondences across viewpoints.}
  A moving bus is observed from different viewpoints at $t=0$, $t=40$, and $t=80$.
  Static points \textcolor[HTML]{E94241}{\textbf{P1}}, \textcolor[HTML]{2FBC5C}{\textbf{P2}}, and \textcolor[HTML]{3677EF}{\textbf{P3}} project to different image locations but retain their world-space coordinates and encoded colors.
  By contrast, the \textcolor[HTML]{8C3018}{\textbf{highlighted region}} on the moving bus changes its world-space position and therefore its XYZ color over time.
  We consequently introduce a separate, time-invariant foreground identity map to preserve correspondence, while the foreground XYZ map represents its time-varying position.}
  \label{fig:world_space_xyz_correspondence}
  \vspace{-12pt}
\end{figure*}

\begin{itemize}[leftmargin=*, itemsep=2pt, topsep=-2pt]
\item \textbf{Preserving identity and 3D position.}
We encode world-space XYZ coordinates as RGB values. Static background coordinates provide position and persistent identity. For moving foreground points, we separate current XYZ from fixed colors identifying handles across frames (Fig.~\ref{fig:world_space_xyz_correspondence}).

\item \textbf{Rigid motion handles.}
Multiple independently controlled rigid handles approximate non-rigid motion---for example, moving a person's hands while keeping the torso still. This provides local control without authoring a dense deformation field or requiring an object template, skeleton, or category-specific motion prior.
\end{itemize}

\vspace{-8pt}
\paragraph{Background and foreground XYZ maps
($\mathbf{F}_t^{\mathrm{bg}}$, $\mathbf{F}_t^{\mathrm{fg}}$).}
We project background and motion-handle points into each frame:
\begin{equation}
\mathbf{u}_t=\pi\!\left(
\mathbf{K}_t(\mathbf{R}_t\mathbf{p}_t+\mathbf{t}_t)
\right),
\label{eq:projection}
\end{equation}
\looseness=-1 where $\pi((a,b,c)^\top)=(a/c,b/c)^\top$
denotes perspective division. Background points remain at
$\mathbf{p}_t=\mathbf{p}_{\mathbf{u}}$, while handle points
$\mathbf{p}_t\in\mathcal{H}_{t,i}$ follow their authored motion.
We resolve visibility by depth and write the visible points into image
maps, using $\mathbf{u}_t$ as the image location and normalized
world-space XYZ as the RGB value. These colors describe geometry,
not the points' appearance.

Both maps share a per-axis normalization fixed throughout the clip.
Let $\mathbf{p}_{\min}$ and $\mathbf{p}_{\max}$ be the coordinate-wise
extrema of the valid first-frame points after depth-outlier handling
(supplementary Sec.~\ref{supp:maps}), with bounding-box center
$\mathbf{o}=(\mathbf{p}_{\max}+\mathbf{p}_{\min})/2$ and half-range
$\boldsymbol{\sigma}=(\mathbf{p}_{\max}-\mathbf{p}_{\min})/2$.
We normalize world coordinates as
\begin{equation}
\mathcal{N}(\mathbf{p})=\frac{1}{2}\left[
\frac{\mathbf{p}-\mathbf{o}}{\boldsymbol{\sigma}}+\mathbf{1}\right],
\qquad \text{where }\mathbf{1}=(1,1,1)^\top.
\label{eq:xyz-normalization}
\end{equation}
For points within the bounding box, $(\mathbf{p}-\mathbf{o})/\boldsymbol{\sigma}$ lies in $[-1,1]^3$, and the shift and scaling map it to $[0,1]^3$. The background and foreground RGB values are
\begin{equation}
\mathbf{c}_t^{\mathrm{bg}}=\mathcal{N}(\mathbf{p}_{\mathbf{u}}),
\qquad
\mathbf{c}_t^{\mathrm{fg}}=\mathcal{N}(\mathbf{p}_t),
\quad \mathbf{p}_t\in\mathcal{H}_{t,i}.
\label{eq:xyz-values}
\end{equation}
A static background point keeps its world position and XYZ color as
the camera moves; only its image location changes. The background map
therefore indicates both \emph{where} the point projects and \emph{which}
world point it represents. A foreground control point's XYZ color
changes as it moves through the world. Using the same
$(\mathbf{o},\boldsymbol{\sigma})$ for both maps expresses foreground
and background positions in a common coordinate frame and scale.

\vspace{-8pt}
\paragraph{Foreground identity maps ($\mathbf{F}_t^{\mathrm{id}}$).}
The foreground XYZ map alone cannot establish persistent correspondence because a moving point changes its world position and encoded color. We therefore assign each handle a time-invariant identity and render it separately.

To represent each handle as a compact image region, we approximate $\mathcal{H}_{t,i}$ by a 3D Gaussian $\mathcal{G}_{t,i}=(\boldsymbol{\mu}_{t,i},\boldsymbol{\Sigma}_{t,i})$, where $\boldsymbol{\mu}_{t,i}$ and $\boldsymbol{\Sigma}_{t,i}$ are the empirical mean and covariance of its points. Projection through Eq.~\ref{eq:projection} yields a 2D Gaussian with mean $\boldsymbol{\nu}_{t,i}$ and covariance $\boldsymbol{\Lambda}_{t,i}$. Let $\Omega_{t,i}$ denote its image-plane footprint.

When handle $i$ is created, we sample a color $\mathbf{c}_i\in[0,1]^3$ and reuse it in every frame. Let $\alpha_{t,i}(\mathbf{u}_t)\in[0,1]$ be the Gaussian weight, largest at the projected center and decreasing toward the boundary. Within the footprint, we render
\begin{equation}
\mathbf{c}_t^{\mathrm{id}}=\alpha_{t,i}(\mathbf{u}_t)\mathbf{c}_i,
\qquad \mathbf{u}_t\in\Omega_{t,i},
\label{eq:fg-map}
\end{equation}
Pixels outside all handle footprints are zero. Where footprints overlap, depth ordering selects the visible handle. The foreground XYZ and identity maps share this footprint, validity mask, and visible handle. The fixed base color $\mathbf{c}_i$ identifies the same handle across frames, while $\alpha_{t,i}$ determines its spatial weight. This construction produces three guidance maps
$\mathbf{F}_t^{\mathrm{bg}},\mathbf{F}_t^{\mathrm{fg}},
\mathbf{F}_t^{\mathrm{id}}\in\mathbb{R}^{H\times W\times3}$,
with invalid pixels set to zero.

In summary, the maps provide complementary correspondence and motion cues. The background XYZ map supplies a static world reference for camera motion. For the foreground, the identity map identifies the projected handle,
but does not explicitly encode its world-space motion. The foreground
XYZ map supplies the control points' current world positions, allowing
the model to interpret their motion relative to the background.

\vspace{-5pt}
\subsection{Architecture}
\label{sec:method:architecture}

We build \method on a pretrained Wan video diffusion model and freeze its VAE and base diffusion weights. We train only a lightweight side branch for geometric guidance, convolutional projections between branches, and LoRA adapters in the main branch.

The frozen VAE encodes all three guidance-map sequences. The main branch receives the noisy video latent, the reference-frame latent, and the encoded foreground identity sequence. The side branch receives all three encoded maps and binary masks marking valid background projections ($\mathbf{M}^{\mathrm{bg}}$), projected handle pixels ($\mathbf{M}^{\mathrm{fg}}$), and the first-frame dynamic region ($\mathbf{M}^{\mathrm{dyn}}$). It operates at half the main branch's spatial resolution and injects features through zero-initialized residual projections. Supplementary Sec.~\ref{supp:architecture} details mask processing, input layouts, and feature transfer.

During training, each guidance sequence describes the motion in its paired
target video. We apply Wan's flow-matching objective to the target video's
noised latents, conditioned on the reference frame, text, and guidance
(supplementary Eq.~\ref{eq:supp:flow-matching}).
The VAE supplies the latent representation; the trainable diffusion
adaptation modules learn how the controls relate to the observed video.

\vspace{-5pt}
\subsection{Sampling Motion Handles for Training}
\label{sec:method:training}

Training videos contain observed motion but no artist-authored handles.
For real clips, we estimate depth and camera trajectories with
ViPE~\citep{huang2025vipe} and obtain motion trajectories with
CoWTracker~\citep{lai2026cowtracker}. Taking the first frame as
input, we use its dynamic mask to sample spatially separated handles
from visible foreground points with valid geometry. Each handle groups
nearby tracks within a scene-scaled 3D ball and follows their motion
through the clip.

After filtering unreliable tracks and handles, we convert the
accepted controls into foreground XYZ and identity maps. We vary the
number and placement of handles across configurations to simulate
partial artist control. Clips without valid foreground trajectories
receive background-only conditioning. Supplementary
Sec.~\ref{supp:handle-sampling} provides the sampling and filtering
details.

\section{Experiments}
\label{sec:Experiment}

\paragraph{Training Setup.}
We build \method upon Wan2.1-Fun-V1.1-14B-Control~\citep{wan2025wan}. We only train a small set of adaptation modules to learn the proposed geometry guidance maps.
The model is trained for approximately $20$k iterations on $4$ H100 GPUs.

\vspace{-8pt}
\paragraph{Datasets.}
We train on $45$k video clips covering diverse camera and object
motions. Approximately $5$k clips come from synthetic datasets,
including Dynamic Replica~\citep{dynamicreplica} and
PointOdyssey~\citep{pointodyssey}, which provide ground-truth geometry
and motion trajectories. The remaining clips come from real-world
datasets, including SpatialVid-HQ~\citep{spatialvidhq}. For these clips,
we estimate depth, camera poses, and dynamic masks using our geometry
extraction pipeline and obtain motion trajectories with
CoWTracker~\citep{lai2026cowtracker}.
All clips are processed and cropped to $480\times832$ resolution.
We generate the missing video captions with Qwen2.5-VL-72B~\citep{bai2025qwen25vltechnicalreport}
and encode them with umT5~\citep{umt5}.

\vspace{-8pt}
\paragraph{Evaluation Setup.}
We evaluate on $75$ examples from the DAVIS dataset and the SpatialVid
test set, grouped by content into Camera Shooting and Object Interactions.
Ground-truth videos and camera poses serve as references.
We measure reference fidelity using full-frame PSNR and LPIPS~\citep{lpips}
and their foreground-region counterparts (R-PSNR and R-LPIPS).
Following CameraCtrl~\citep{he2024cameractrl}, we report camera
rotation (Rot) and translation (Trans) errors.
We assess video quality using the subject consistency, background consistency,
and imaging quality metrics from VBench-I2V~\citep{huang2024vbench}.

\vspace{-8pt}
\paragraph{Baselines and Control Inputs.}
We compare with Wan2.1~\citep{wan2025wan} and five controllable video generation methods.
In our evaluation, ATI~\citep{wang2025ati}, Wan-Move~\citep{chu2025wanmove},
and Go-with-the-Track~\citep{namekata2026gowiththetrack} receive dense 2D
foreground and background trajectories extracted from ground-truth videos.
VerseCrafter~\citep{zheng2026versecrafter} and SymphoMotion~\citep{zhang2026symphomotion}
provide camera control and coarse object-level motion control.
\method uses sparse 3D motion handles and a target camera trajectory
derived from the same reference motion.
Dense ground-truth trajectories are unavailable when authoring new motion from
a single image; a similar input asymmetry is discussed in concurrent work~\citep{liu2026moright}.

\vspace{-5pt}
\subsection{Main Results}
\label{sec:MainResults}

\begin{figure*}[tp]
  \centering
  \includegraphics[width=\textwidth,height=\dimexpr\textheight-70pt\relax,keepaspectratio]{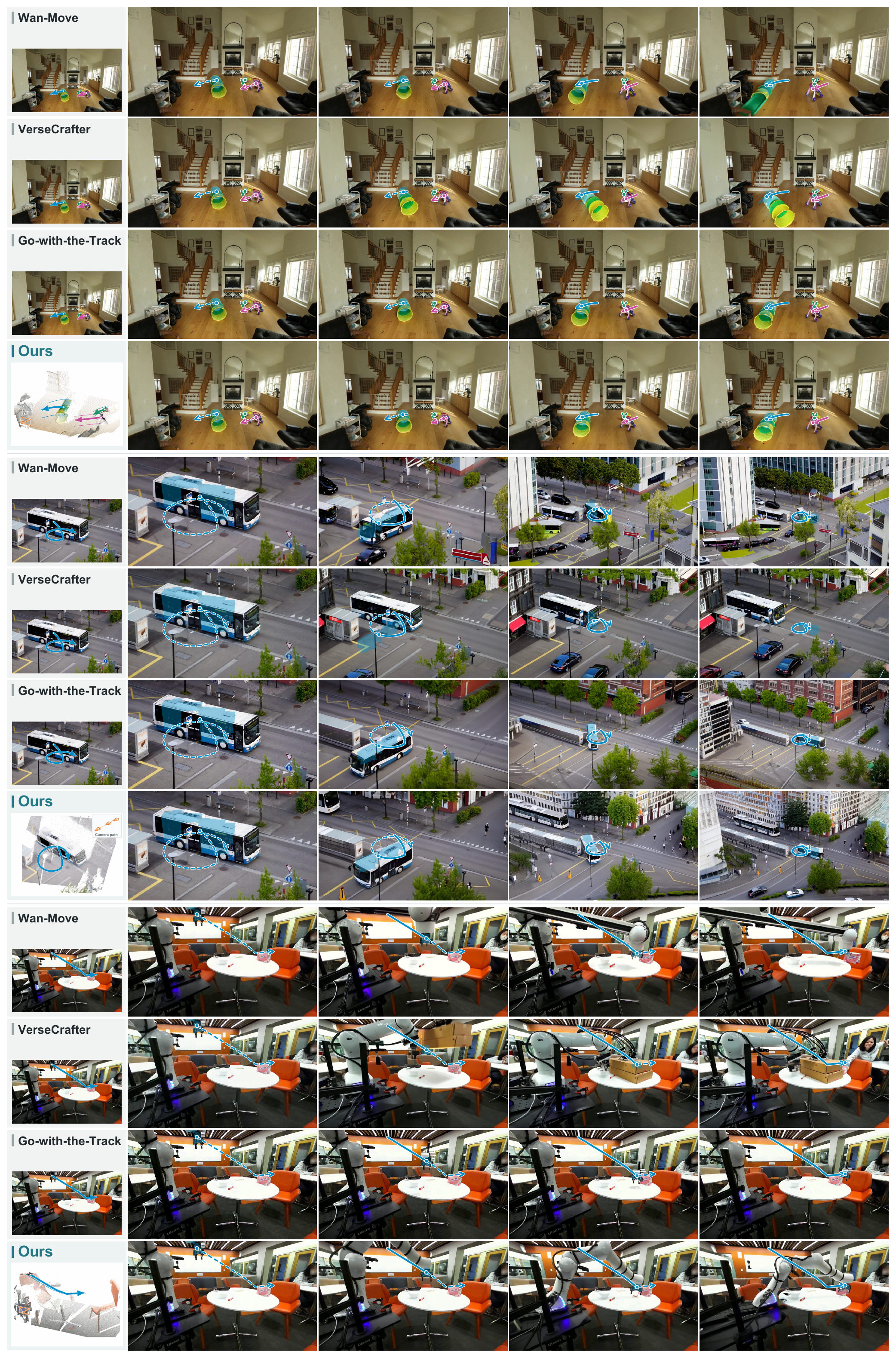}
  \caption{
  \textbf{Qualitative comparison on artist-authored motion.}
  Living room, Bus, and Robotic arm (top to bottom) each show
  Wan-Move, VerseCrafter, Go-with-the-Track, and \method (ours), in row order.
  The first column shows 3D controls (\method) or shared 2D projections (baselines);
  four output frames sample each control interval.
  Solid/dashed paths show elapsed/remaining target motion, dots mark current
  targets, and translucent masks show projected target regions.
  }
  \label{fig:qualitative_comparison}
\end{figure*}

Fig.~\ref{fig:qualitative_comparison} compares \method with three baselines
on artist-authored examples.
In Living room, \method moves the green tunnel with less distortion than
Wan-Move and VerseCrafter.
In Bus, \method coordinates the intended turn with camera pullback,
while VerseCrafter largely retains the bus's initial orientation.
Robotic arm further illustrates motion dependencies between connected
parts. Go-with-the-Track moves the gripper region but leaves most of the
arm static, while VerseCrafter introduces extraneous boxes.
\method follows the end-effector path while articulating the connected
arm segments. This coordinated response suggests that \method can infer
motion dependencies within an articulated structure, using sparse local
guidance to generate motion beyond the directly controlled region.

\begin{table*}[t!]
  \caption{
  Quantitative comparison on examples from Camera Shooting and Object Interactions.
  Metrics prefixed with R- are computed within the controlled foreground region.
  Subj., Bkg., and IQ denote VBench subject consistency, background consistency,
  and imaging quality, respectively.
  }
  \vspace{-0.18em}
  \label{tab:quantitative_results}
  \centering
  \small
  \setlength{\tabcolsep}{3pt}
  \resizebox{\linewidth}{!}{%
  \begin{tabular}{l cc cccc ccc cc cccc ccc}
    \toprule
    & \multicolumn{9}{c}{Camera Shooting}
    & \multicolumn{9}{c}{Object Interactions} \\
    \cmidrule(lr){2-10} \cmidrule(lr){11-19}
    & \multicolumn{2}{c}{Camera Traj.}
    & \multicolumn{4}{c}{Quality}
    & \multicolumn{3}{c}{Consistency}
    & \multicolumn{2}{c}{Camera Traj.}
    & \multicolumn{4}{c}{Quality}
    & \multicolumn{3}{c}{Consistency} \\
    \cmidrule(lr){2-3} \cmidrule(lr){4-7} \cmidrule(lr){8-10}
    \cmidrule(lr){11-12} \cmidrule(lr){13-16} \cmidrule(lr){17-19}
    Method
    & Rot$\downarrow$ & Trans$\downarrow$
    & PSNR$\uparrow$ & LPIPS$\downarrow$ & R-PSNR$\uparrow$ & R-LPIPS$\downarrow$
    & Subj.$\uparrow$ & Bkg.$\uparrow$ & IQ$\uparrow$
    & Rot$\downarrow$ & Trans$\downarrow$
    & PSNR$\uparrow$ & LPIPS$\downarrow$ & R-PSNR$\uparrow$ & R-LPIPS$\downarrow$
    & Subj.$\uparrow$ & Bkg.$\uparrow$ & IQ$\uparrow$ \\
    \midrule
Wan2.1
& 4.98 & 2.40 & 12.44 & 0.49 & 14.08 & 0.34 & 0.87 & 0.92 & 0.66
& 5.14 & 2.53 & 13.08 & 0.47 & 12.11 & 0.38 & 0.87 & \underline{0.92} & \underline{0.67} \\

ATI
& 3.80 & 1.15 & 16.18 & \underline{0.38} & 16.66 & 0.20 & 0.88 & 0.92 & 0.67
& 3.97 & 1.26 & 15.87 & 0.39 & 14.29 & 0.25 & \underline{0.88} & 0.91 & 0.66 \\

Wan-Move
& 3.73 & \underline{1.02} & 16.58 & \textbf{0.36} & \underline{17.13} & \underline{0.19} & \underline{0.90} & \underline{0.93} & \textbf{0.69}
& 3.76 & \underline{1.08} & 16.51 & \underline{0.37} & \textbf{15.46} & \underline{0.22} & \textbf{0.89} & \textbf{0.94} & \underline{0.67} \\

VerseCrafter
& 3.92 & 1.27 & 15.27 & 0.41 & 15.82 & 0.27 & \underline{0.90} & 0.92 & \textbf{0.69}
& 4.02 & 1.24 & 13.94 & 0.43 & 13.27 & 0.33 & \textbf{0.89} & \underline{0.92} & \textbf{0.68} \\

SymphoMotion
& 5.04 & 2.45 & 12.29 & 0.50 & 13.94 & 0.35 & 0.86 & 0.91 & 0.65
& 5.20 & 2.58 & 12.91 & 0.48 & 11.96 & 0.39 & 0.86 & \underline{0.92} & 0.66 \\

Go-w-Track
& \textbf{3.69} & 1.07
& \underline{16.61} & \textbf{0.36}
& 17.11 & \underline{0.19}
& \textbf{0.91} & \textbf{0.94} & \textbf{0.69}
& \textbf{3.73} & \underline{1.08}
& \textbf{16.59} & \underline{0.37}
& \underline{15.43} & \underline{0.22}
& \underline{0.88} & \underline{0.92} & \underline{0.67} \\

\method (ours)
& \underline{3.70} & \textbf{1.00}
& \textbf{16.64} & \textbf{0.36}
& \textbf{17.16} & \textbf{0.18}
& \underline{0.90} & \textbf{0.94} & \underline{0.68}
& \underline{3.74} & \textbf{1.05}
& \underline{16.54} & \textbf{0.35}
& 15.40 & \textbf{0.21}
& \textbf{0.89} & \underline{0.92} & \underline{0.67} \\
    \bottomrule
  \end{tabular}%
  }
  \vspace{-10pt}
\end{table*}

In Tab.~\ref{tab:quantitative_results}, \method achieves the lowest
translation error and R-LPIPS in both categories, ranking second in
rotation error behind Go-with-the-Track.
On Camera Shooting, it leads PSNR ($16.64$) and R-PSNR ($17.16$), and
ties Wan-Move and Go-with-the-Track for lowest LPIPS ($0.36$ at reported precision).
On Object Interactions, it achieves the lowest LPIPS ($0.35$), with
PSNR ($16.54$) second to Go-with-the-Track ($16.59$).
Other methods retain complementary strengths.
Go-with-the-Track leads subject consistency on Camera Shooting;
on Object Interactions, VerseCrafter leads imaging quality, while
Wan-Move leads R-PSNR and background consistency.
\method's clearest advantages are translation accuracy and foreground
perceptual fidelity.

These results indicate effective camera and foreground control with sparse
3D motion handles and a camera trajectory.
Additional comparison examples and video versions of these results
are available on the project website.

\vspace{-5pt}
\subsection{Ablation Study}
\label{sec:Ablation}

We ablate the background XYZ map $\mathbf{F}^{\mathrm{bg}}$, the foreground
XYZ map $\mathbf{F}^{\mathrm{fg}}$, and the side-branch spatial downsampling
factor $s_{\mathrm{side}}$ in \method.
The default retains all guidance maps with $s_{\mathrm{side}}=2$
(half the main branch's spatial resolution).
Quantitative variants share training data, backbone, and optimization settings,
differing only in the indicated component.
Tab.~\ref{tab:quantitative_results2_2} reports results on Camera Shooting examples.

\begin{wraptable}[7]{r}{0.49\textwidth}
  \centering
  \footnotesize
  \vspace{-2mm}
  \caption{Guidance-map and side-branch resolution ablations on Camera Shooting examples.}
  \vspace{-2mm}
  \label{tab:quantitative_results2_2}
  \resizebox{\linewidth}{!}{%
  \begin{tabular}{l c c c c c}
    \toprule
    Method
    & R-LPIPS$\downarrow$
    & R-PSNR$\uparrow$
    & Subj. $\uparrow$
    & Bkg. $\uparrow$
    & IQ $\uparrow$ \\
    \midrule
    \method (w/o $\mathbf{F}^{\mathrm{bg}}$)    & 0.235 & 16.20 & 0.889 & 0.937 & 0.683 \\
    \method (w/o $\mathbf{F}^{\mathrm{fg}}$)    & 0.223 & 15.26 & 0.912 & 0.938 & 0.679 \\
    \method ($s_{\mathrm{side}}=1$) & 0.189 & 17.17 & 0.897 & 0.942 & 0.686 \\
    \method ($s_{\mathrm{side}}=2$) & 0.184 & 17.16 & 0.902 & 0.944 & 0.691 \\
    \bottomrule
  \end{tabular}
  }
  \vspace{2mm}
\end{wraptable}

Removing $\mathbf{F}^{\mathrm{bg}}$ degrades foreground reference fidelity:
R-LPIPS increases from $0.184$ to $0.235$, and R-PSNR drops from $17.16$ to $16.20$.
These results support using background correspondence as a reference for foreground control.

\begin{figure*}[t]
  \centering
  \includegraphics[width=\textwidth]{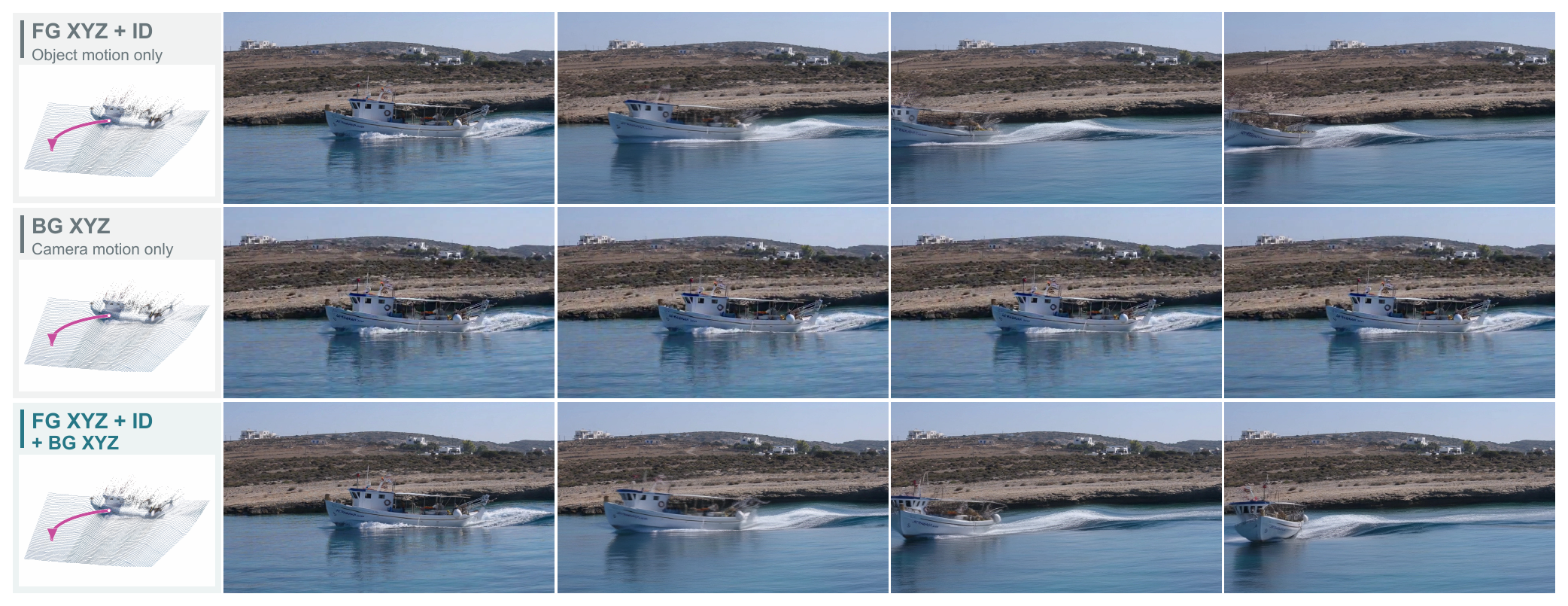}
  \vspace{-15pt}
  \caption{\textbf{Control map ablation for \method.}
  The target is a fixed camera with the boat following the purple 3D trajectory.
  Foreground XYZ and identity maps alone (top) show background drift.
  Background XYZ alone (middle) provides no explicit object motion control.
  Full guidance (bottom) combines all three maps, better preserving the static
  background while guiding the boat.
  }
  \label{fig:boat_ablation}
  \vspace{-15pt}
\end{figure*}

As a qualitative illustration, Fig.~\ref{fig:boat_ablation} shows the intended task
of moving the boat while keeping the camera fixed. The foreground-only result
exhibits unwanted background motion, while the result with full foreground
and background XYZ guidance better preserves the static scene. The middle
row additionally shows background XYZ guidance only. The saved camera exports
differ between the foreground-only and full-guidance runs, which limits
attribution to background guidance alone.

Removing $\mathbf{F}^{\mathrm{fg}}$ increases R-LPIPS to $0.223$
and reduces R-PSNR to $15.26$.
Despite the highest subject consistency ($0.912$), this variant reconstructs
foreground motion less faithfully.
With all maps retained, $s_{\mathrm{side}}=2$ improves R-LPIPS from $0.189$ to
$0.184$ relative to $s_{\mathrm{side}}=1$, while R-PSNR decreases slightly from
$17.17$ to $17.16$.
It also improves subject consistency ($0.897$ to $0.902$),
background consistency ($0.942$ to $0.944$), and imaging quality ($0.686$ to $0.691$).

Overall, the default configuration achieves the lowest R-LPIPS and the highest
background consistency and imaging quality among the tested variants.
Additional qualitative examples appear in Figs.~\ref{fig:teaser}, \ref{fig:3d_control_teaser_example}, and~\ref{fig:method_comparison_results}.

\vspace{-5pt}

\section{Discussion}
\label{sec:discussion}
We presented \method, a system for jointly
authoring camera and object motion in a shared 3D world lifted from a
single image. Its guidance maps encode static background coordinates,
persistent foreground handle identities, and current world-space
positions in a form compatible with pretrained video diffusion models.
This representation makes the intended 3D motion explicit where
image-space trajectories alone leave it ambiguous. Our results show
that a few local handles can compose camera and foreground motion,
including a piecewise-rigid approximation to articulated motion,
through lightweight adaptation of a pretrained video generator.

The single-image scaffold leaves occluded regions unmodeled, requiring
the generator to synthesize unseen surfaces under large viewpoint
changes. Local rigid handles are better suited to articulated motion
than to fracture, tearing, or fluid dynamics, and authoring many handles
can become cumbersome for distributed deformation. Independently
authored handles also impose no physical constraints or causal
interactions between objects. Geometry completion, automatic handle
placement, and learned interaction priors could extend the range of
shots artists can compose.

\clearpage

\section*{Acknowledgments}
We thank Shaoheng Fang and Jiahao Wang for helpful discussions and suggestions about the training data construction and result evaluation.%

\section*{AI Use Statement}
We used generative AI only for language polishing in manuscript preparation.
Qwen2.5-VL-72B generated missing training captions
(Sec.~\ref{sec:Experiment}). The authors take responsibility for all content.

\section*{Ethics Statement}
Generated videos should be identified
as synthetic to reduce misleading use, with privacy and applicable data
and model licenses respected.

\section*{Reproducibility Statement}
Sections~\ref{sec:method} and~\ref{sec:Experiment} describe the method,
datasets, training setup, and evaluation. The supplement provides
implementation details (Secs.~\ref{supp:scaffold}--\ref{supp:inference}).

\clearpage

\appendix
\section*{Supplementary Material}

Project website and videos: \url{https://generative-cinematographer.github.io/}.

This supplement begins with additional qualitative comparisons
(Sec.~\ref{supp:qualitative-comparisons}).
It then reviews latent video diffusion (Sec.~\ref{supp:diffusion}) and details
the scene scaffold (Sec.~\ref{supp:scaffold}), guidance maps (Sec.~\ref{supp:maps}),
architecture (Sec.~\ref{supp:architecture}), training (Sec.~\ref{supp:training}),
and inference (Sec.~\ref{supp:inference}). We conclude with failure cases and
limitations (Sec.~\ref{supp:failures}).

\section{Additional Qualitative Comparisons}
\label{supp:qualitative-comparisons}

\begin{figure*}[!htbp]
  \centering
  \newcommand{\ctframe}[1]{\includegraphics[width=0.15\textwidth]{#1}}
  \newcommand{\ctlabel}[1]{\footnotesize\rotatebox{90}{\makebox[1.6cm][c]{#1}}}
  \setlength{\tabcolsep}{1.0pt}
  \renewcommand{\arraystretch}{0.84}
  \begin{tabular}{@{}ccccccc@{}}
    & \scriptsize $t_0$
    & \scriptsize $t_1$
    & \scriptsize $t_2$
    & \scriptsize $t_3$
    & \scriptsize $t_4$
    & \scriptsize $t_5$ \\
    \ctlabel{Wan-Move}
    & \ctframe{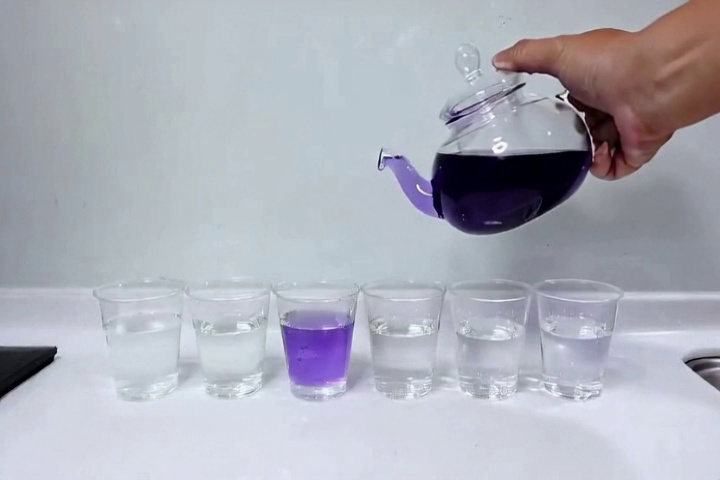}
    & \ctframe{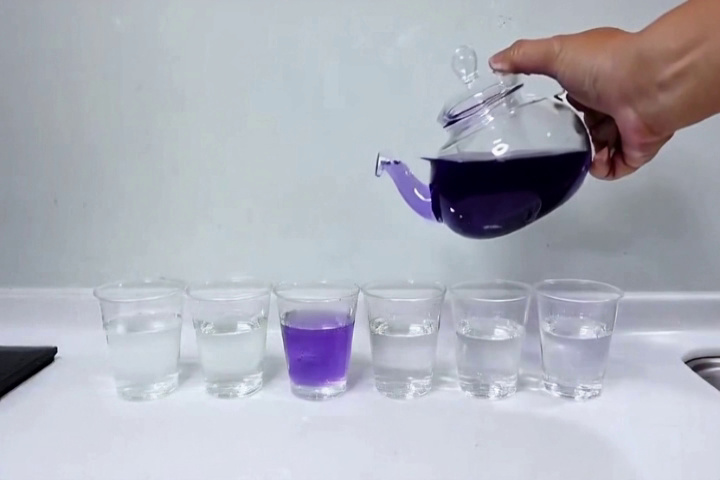}
    & \ctframe{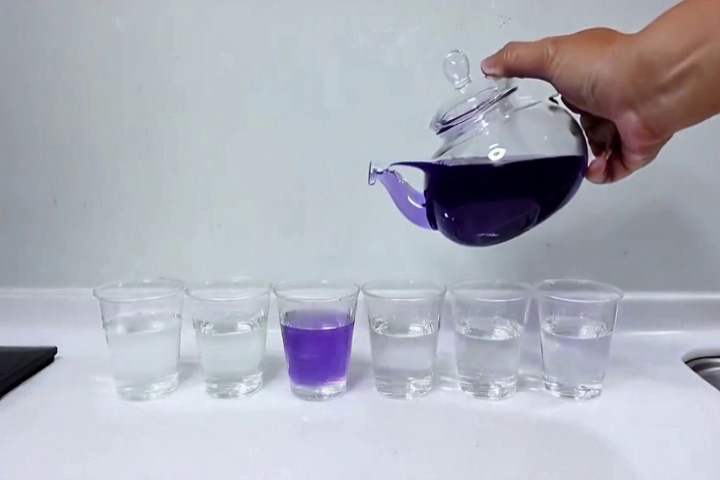}
    & \ctframe{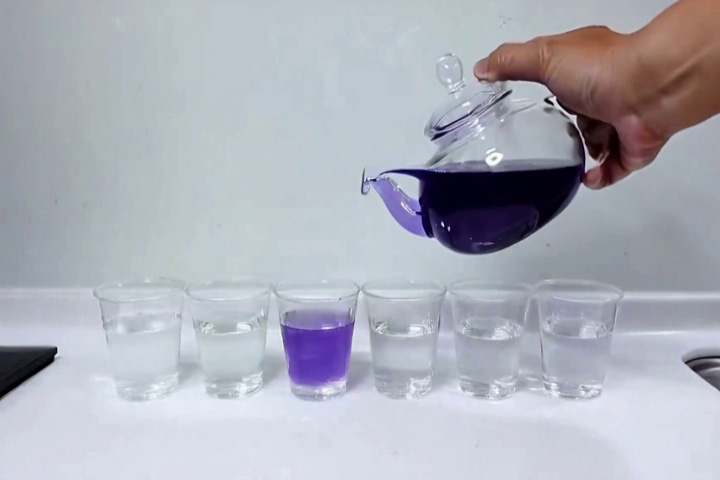}
    & \ctframe{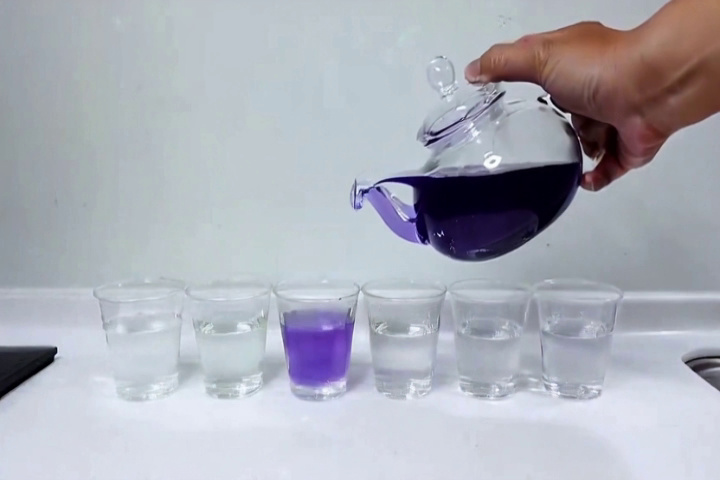}
    & \ctframe{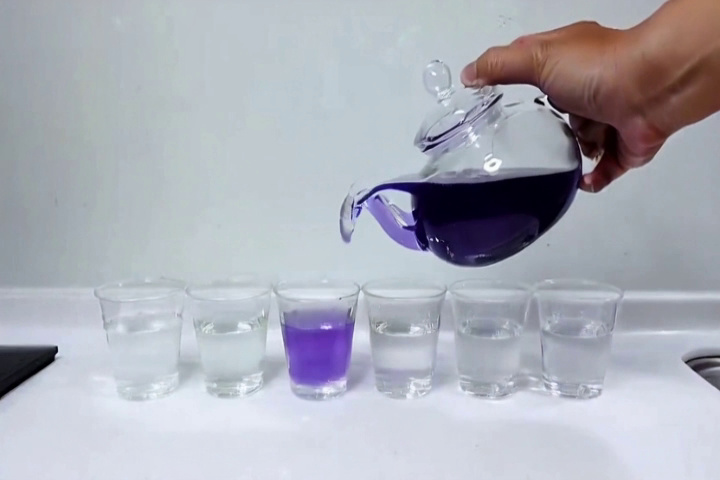} \\
    \ctlabel{\method}
    & \ctframe{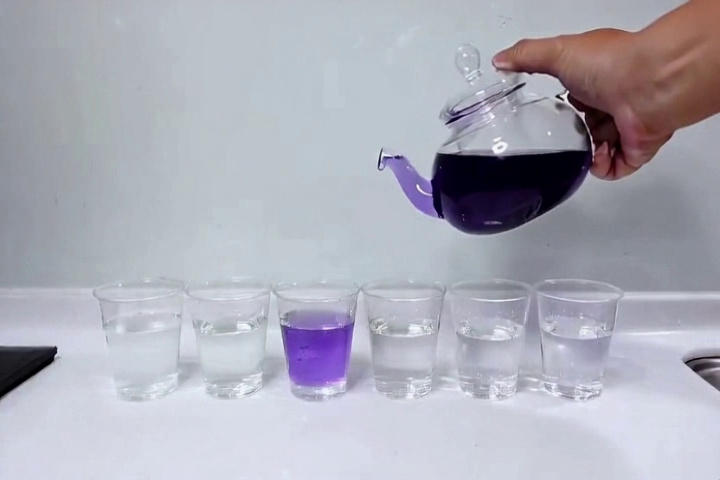}
    & \ctframe{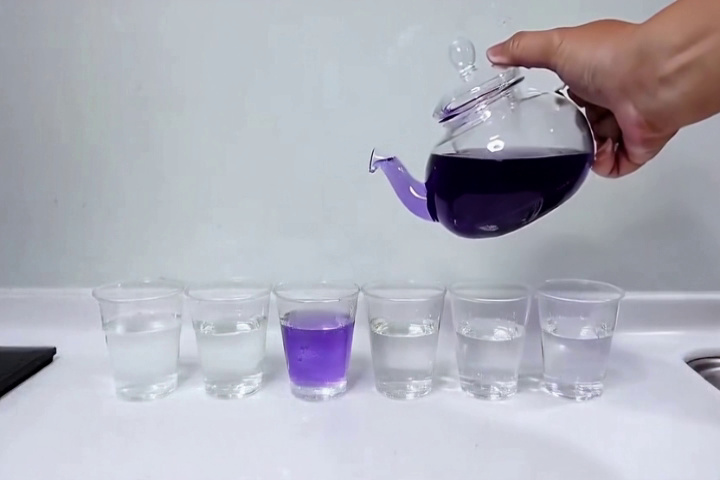}
    & \ctframe{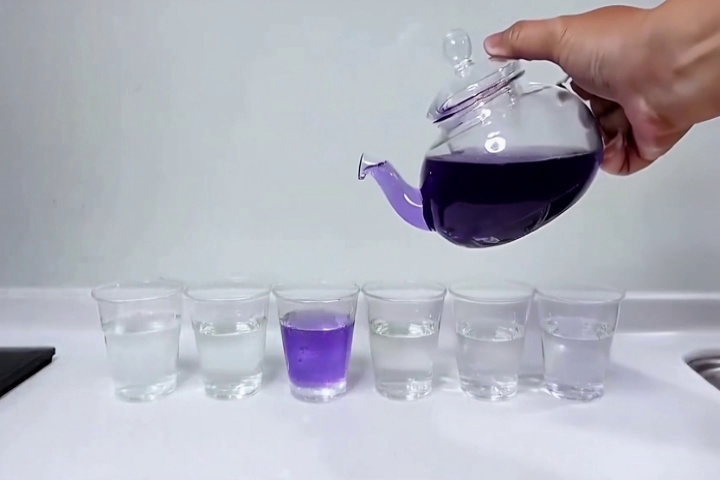}
    & \ctframe{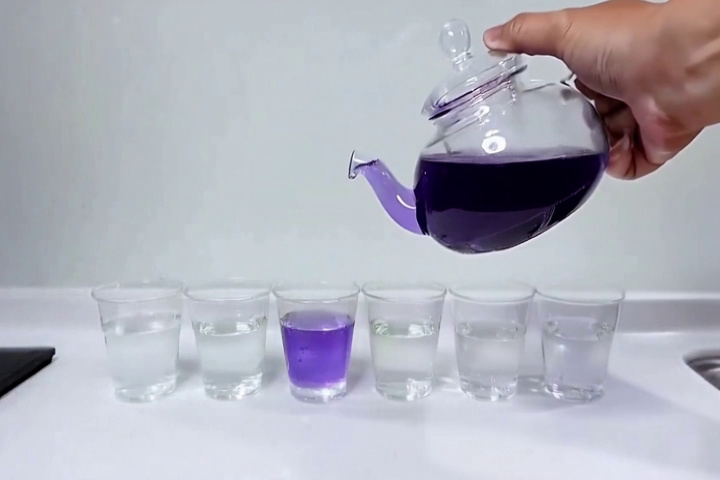}
    & \ctframe{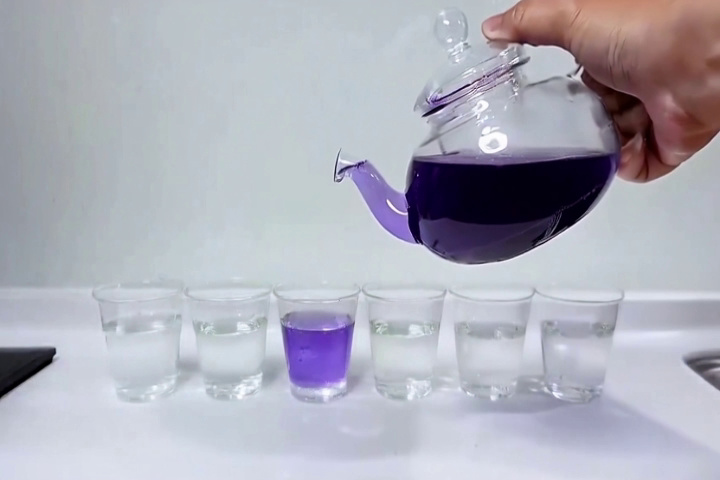}
    & \ctframe{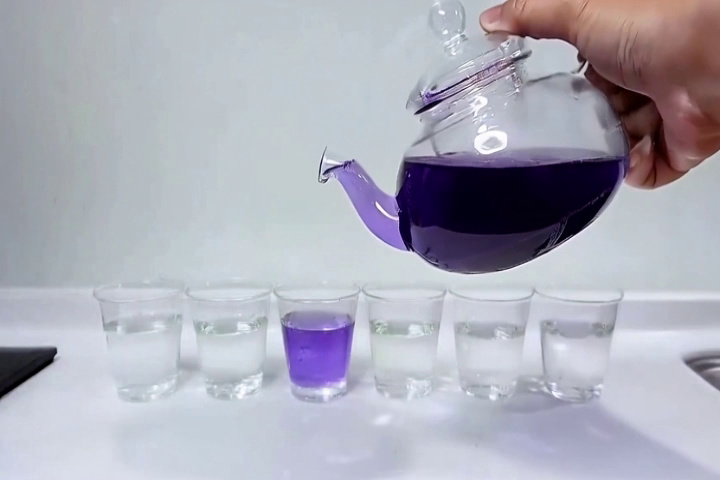}
  \end{tabular}
  \vspace{-0.75em}
  \caption{Comparison on an ambiguous motion-control example where the intended edit moves the teapot toward the camera while preserving its orientation in world space. Image-plane trajectories do not uniquely specify the underlying 3D transformation. Wan-Move~\citep{chu2025wanmove} produces geometrically inconsistent motion in this example, while \method better preserves the intended orientation and scene geometry using handles authored in the canonical world frame.}
  \label{fig:3d_control_teaser_example}
\end{figure*}

\paragraph{Ambiguity in image-space control.}
Fig.~\ref{fig:3d_control_teaser_example} examines an edit that moves a teapot
toward the camera while preserving its orientation in world space.
Image-plane trajectories constrain projected motion but do not uniquely
specify the underlying 3D transformation. In this example, Wan-Move produces
geometrically inconsistent motion across the sampled frames. In \method, 3D handles
specify the intended displacement in the canonical world frame, and the
generated sequence better preserves the teapot's orientation and scene geometry.
This example illustrates the benefit of making the intended 3D motion explicit
when different spatial motions can have similar image-plane projections.

\FloatBarrier

\begin{figure*}[!htbp]
  \centering
  \newcommand{\mcframe}[1]{\includegraphics[width=0.12\textwidth]{#1}}
  \newcommand{\mclabel}[2]{\footnotesize\shortstack{#1\\#2}}
  \newcommand{\mcheader}{%
    & \footnotesize $t_0$
    & \footnotesize $t_1$
    & \footnotesize $t_2$
    & \footnotesize $t_3$
    & \footnotesize $t_4$
    & \footnotesize $t_5$ \\[-1pt]}
\newcommand{\mcrow}[3]{%
    \mclabel{}{\rotatebox{90}{\makebox[1.15cm][c]{#2}}}
    & \mcframe{figs/compressed/method_comparison_results/#3/frame_0000.jpg}
    & \mcframe{figs/compressed/method_comparison_results/#3/frame_0002.jpg}
    & \mcframe{figs/compressed/method_comparison_results/#3/frame_0005.jpg}
    & \mcframe{figs/compressed/method_comparison_results/#3/frame_0007.jpg}
    & \mcframe{figs/compressed/method_comparison_results/#3/frame_0010.jpg}
    & \mcframe{figs/compressed/method_comparison_results/#3/frame_0012.jpg}}
  \newcommand{\mcoverlayrow}[2]{%
    \mclabel{#1}{Overlay}
    & \mcframe{figs/generated_video_results/#2/frame_0000.png}
    & \mcframe{figs/generated_video_results/#2/frame_0002.png}
    & \mcframe{figs/generated_video_results/#2/frame_0005.png}
    & \mcframe{figs/generated_video_results/#2/frame_0007.png}
    & \mcframe{figs/generated_video_results/#2/frame_0010.png}
    & \mcframe{figs/generated_video_results/#2/frame_0012.png}}
  \setlength{\tabcolsep}{0.55pt}
  \renewcommand{\arraystretch}{0.62}
  \begin{tabular}{@{}ccccccc@{}}
    \mcheader
    \mcrow{Hike}{ATI}{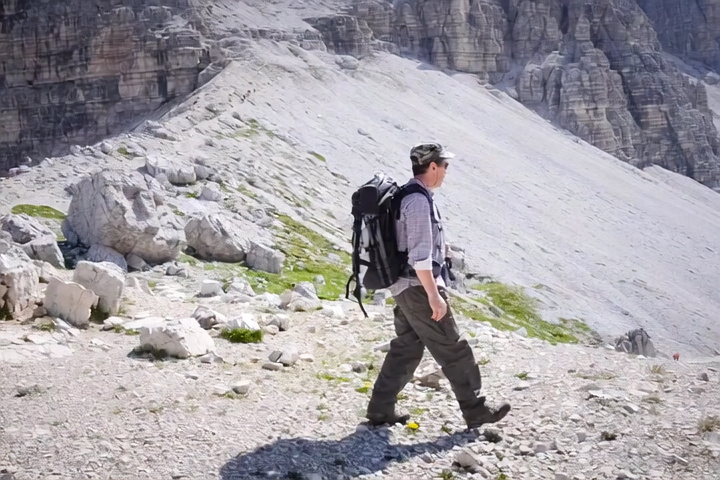} \\
    \mcrow{Hike}{Wan-Move}{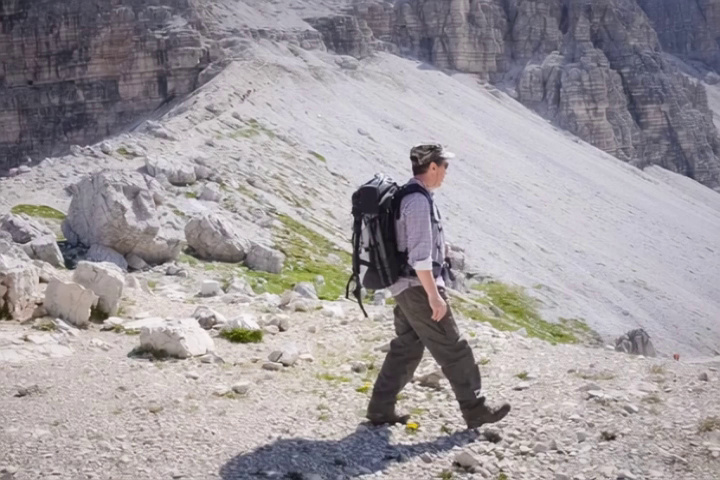} \\
    \mcrow{Hike}{\method}{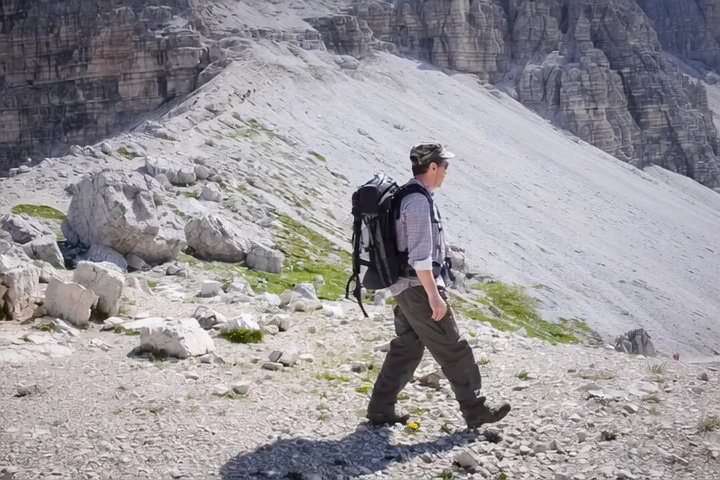} \\
    \mcrow{Hike}{GT}{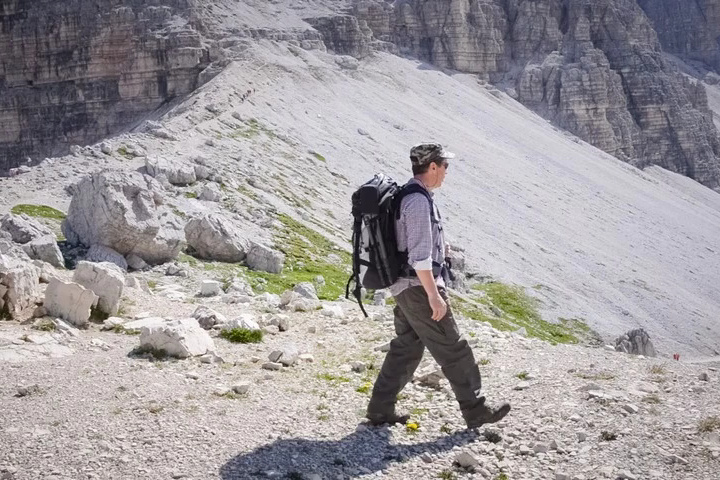} \\
    \addlinespace[0.4pt]
    \mcrow{Camel}{ATI}{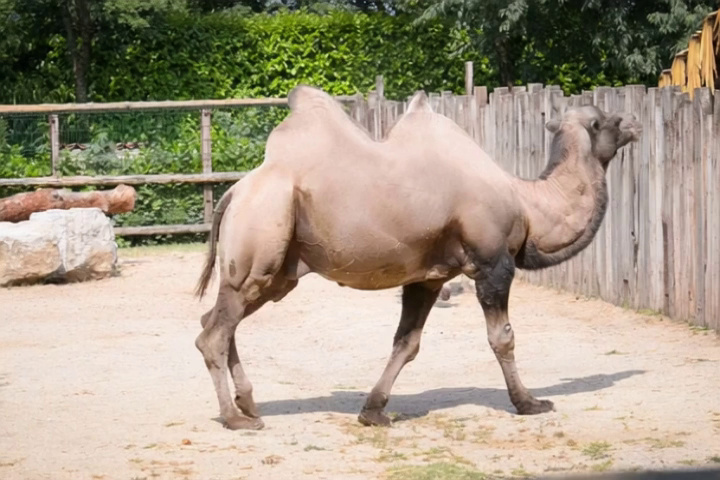} \\
    \mcrow{Camel}{Wan-Move}{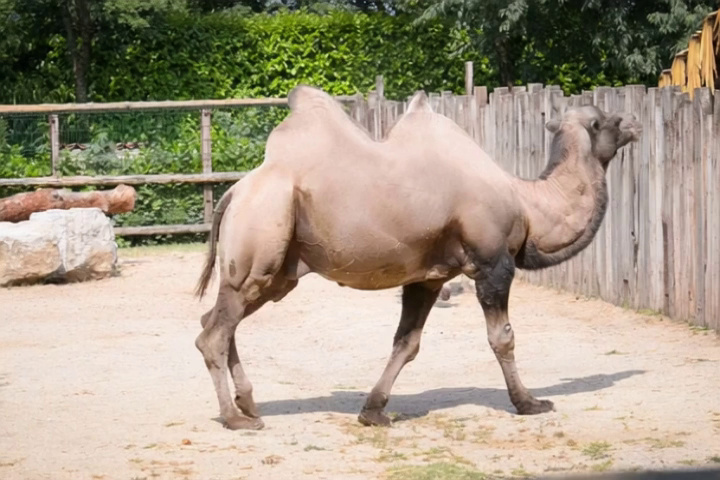} \\
    \mcrow{Camel}{\method}{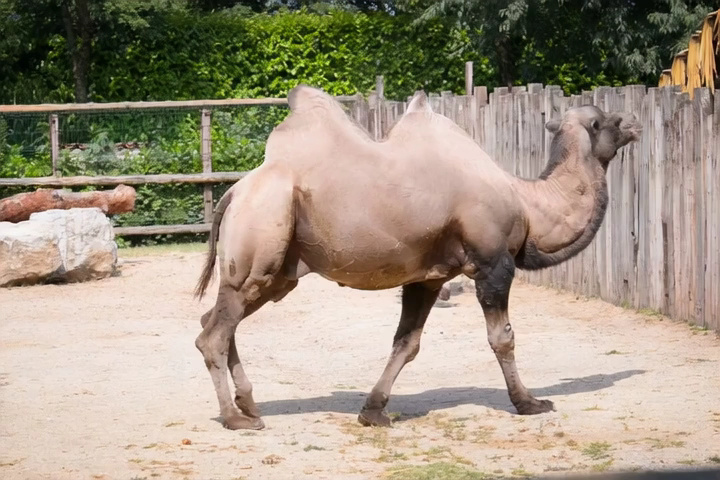} \\
    \mcrow{Camel}{GT}{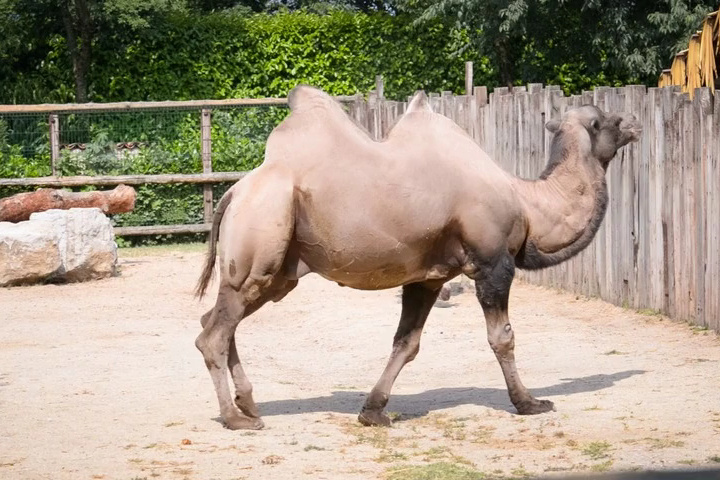} \\
    \addlinespace[0.4pt]
    \mcrow{Dog}{ATI}{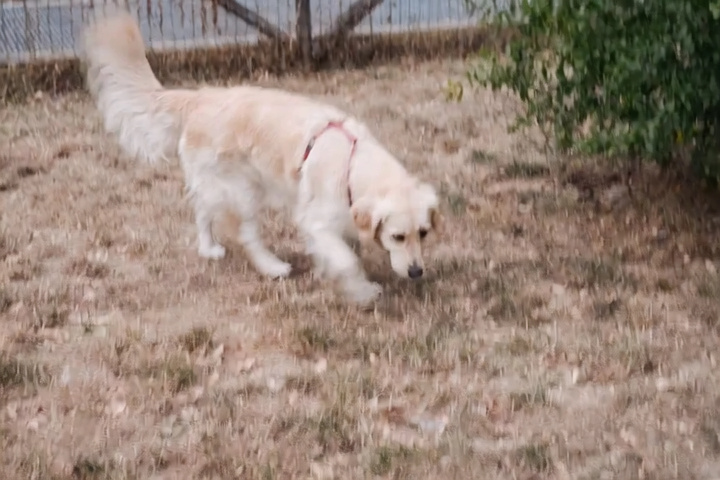} \\
    \mcrow{Dog}{Wan-Move}{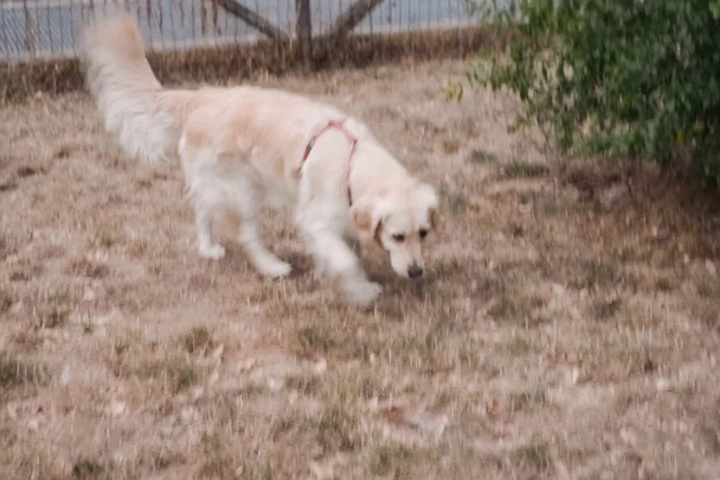} \\
    \mcrow{Dog}{\method}{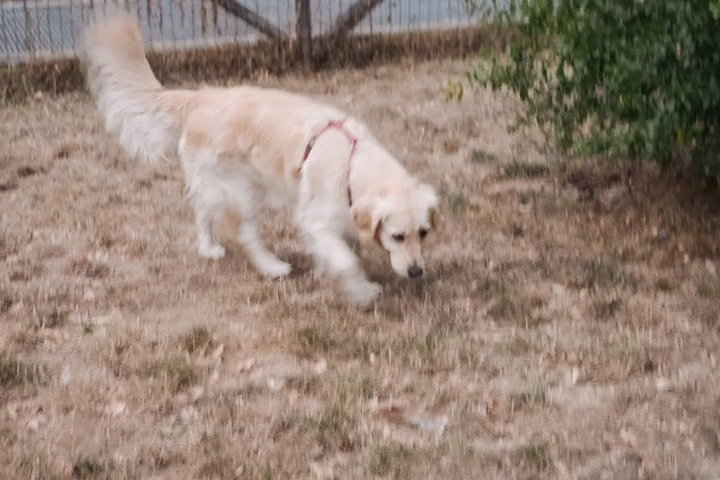} \\
    \mcrow{Dog}{GT}{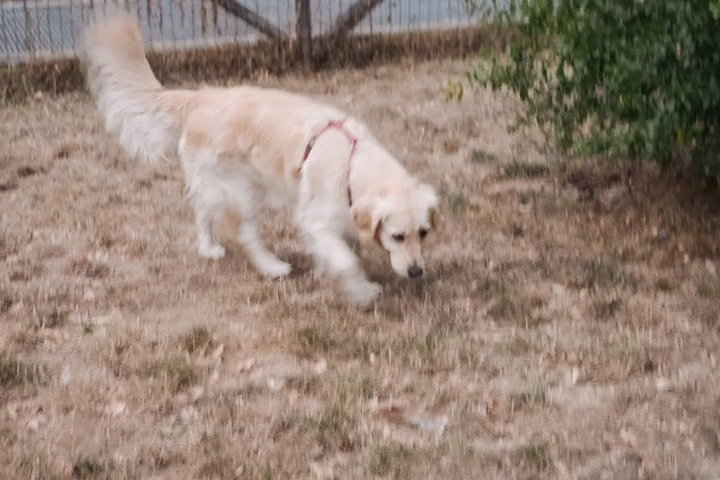} \\
  \end{tabular}
  \vspace{-1.25em}
  \caption{
Qualitative comparison across camera and object motion sequences. Each scene is sampled at six aligned timesteps for ATI~\citep{wang2025ati}, Wan-Move~\citep{chu2025wanmove}, \method, and ground truth. \method follows the reference foreground motion while preserving coherent scene context. In the dog sequence, its generated poses more closely match the reference poses than the baseline outputs.
  }
  \label{fig:method_comparison_results}
\end{figure*}

\paragraph{Camera and object motion.}
Fig.~\ref{fig:method_comparison_results} compares outputs from ATI~\citep{wang2025ati},
Wan-Move~\citep{chu2025wanmove}, and \method with ground-truth videos
for the same foreground-control examples: hiking, camel, and dog sequences.
Each scene shows ATI, Wan-Move, \method, and ground truth in row order,
sampled at six aligned timesteps.
The hiking and camel examples illustrate foreground displacement and changes
in camera-relative appearance, while the dog sequence highlights articulated
motion. \method follows the specified foreground motion while preserving
coherent scene context across the sampled frames. In the dog example, its
generated poses more closely match the reference poses than the baseline
outputs. These results complement the reference-fidelity metrics in
Tab.~\ref{tab:quantitative_results}.

\FloatBarrier

\section{Latent Video Diffusion Background}
\label{supp:diffusion}

Video diffusion models~\citep{ho2020denoising, blattmann2023SVD, wan2025wan} learn a noise-to-video generation process. Latent models reduce computation by operating in a compressed space produced by a pretrained video VAE~\citep{rombach2022high, blattmann2023SVD}. The Wan backbone used by \method adopts a flow-matching objective in this space (Eq.~\ref{eq:supp:flow-matching}).

Given an input video $\mathbf{V} \in \mathbb{R}^{(1+T) \times H \times W \times 3}$, where $1+T$ is the number of frames and $H \times W$ is the spatial resolution, the VAE encoder $\mathcal{E}$ compresses the video temporally by a factor $f_t$, spatially by a factor $f_s$, and expands the channel dimension to $C$:
\begin{equation}
\mathbf{z} = \mathcal{E}(\mathbf{V}) \in \mathbb{R}^{\left(1 + \frac{T}{f_t}\right) \times \frac{H}{f_s} \times \frac{W}{f_s} \times C}.
\label{eq:supp:latent}
\end{equation}
The VAE decoder $\mathcal{D}$ reconstructs the video from the latent: $\hat{\mathbf{V}} = \mathcal{D}(\mathbf{z})$. For Wan2.1~\citep{wan2025wan}, $f_t = 4$, $f_s = 8$, and $C = 16$.

We represent the three guidance maps as RGB sequences so that the same frozen VAE can encode them. Binary masks are processed separately (Sec.~\ref{supp:architecture}). To retain the pretrained visual prior, we freeze the VAE and base diffusion weights and train only the side branch, inter-branch projections, and main-branch LoRA adapters.

\section{Scene Scaffold Details}
\label{supp:scaffold}

\paragraph{Depth and camera estimation.} We estimate first-frame depth with ViPE~\citep{huang2025vipe} and use the initial camera frame as the world frame. The intrinsic matrix $\mathbf{K}_0$ is taken from the depth pipeline's estimate when not provided as input. Pixels with invalid depth are excluded from the initial point cloud. Geometry-supported filling of rasterized background holes is described separately in Sec.~\ref{supp:maps}.

\paragraph{Dynamic mask.} GroundingSAM2 predicts the first-frame dynamic mask $\mathbf{M}^{\mathrm{dyn}} \in \{0,1\}^{H \times W}$, where one marks potentially mobile foreground. This mask partitions the initial point cloud and supplies first-frame conditioning. For the conditioning sequence, we pad subsequent frames with ones; this padding is not a prediction of foreground occupancy in later frames.

\paragraph{Coordinate convention.} The initial camera frame is taken as the canonical world frame, with rotation $\mathbf{R}_0 = \mathbf{I}_3$ and translation $\mathbf{t}_0 = \mathbf{0}$. Subsequent camera poses are expressed as world-to-camera transforms. Per-frame intrinsics $\mathbf{K}_t$ are constant in our experiments unless the artist explicitly authors a zoom; we expose intrinsics per-frame in the construction (Sec.~\ref{sec:method:correspondence} of the main paper) for generality.

\paragraph{Back-projection and point cloud.} Let $\Omega=\{0,\ldots,W-1\}\times\{0,\ldots,H-1\}$ denote the image domain and $\Omega_{\mathrm{valid}}\subseteq\Omega$ the pixels with valid depth. For $\mathbf{u}=(u,v)$, let $\widetilde{\mathbf{u}}=(u,v,1)^\top$ be its homogeneous coordinate. Given the first-frame depth map $\mathbf{D}_0\in\mathbb{R}^{H\times W}$, back-projection gives
\begin{equation}
\mathbf{p}_{\mathbf{u}}=\mathbf{D}_0(\mathbf{u})\mathbf{K}_0^{-1}\widetilde{\mathbf{u}}
=(x_{\mathbf{u}},y_{\mathbf{u}},z_{\mathbf{u}})^\top\in\mathbb{R}^3.
\label{eq:supp:backproj}
\end{equation}
With the input RGB value $\mathbf{c}_{\mathbf{u}}=(R_{\mathbf{u}},G_{\mathbf{u}},B_{\mathbf{u}})^\top$, the colored point cloud is
\begin{equation}
\mathbf{P}=\bigl\{(\mathbf{p}_{\mathbf{u}},\mathbf{c}_{\mathbf{u}})\bigm|\mathbf{u}\in\Omega_{\mathrm{valid}}\bigr\}.
\label{eq:point-cloud}
\end{equation}
The dynamic mask partitions $\mathbf{P}$ into the static background $\mathbf{P}^{\mathrm{bg}}$ and potentially mobile foreground $\mathbf{P}^{\mathrm{fg}}$. Throughout the construction, $\mathbf{u}$ denotes a 2D pixel coordinate, $\mathbf{p}$ a 3D world-space point, and $\mathbf{c}$ an RGB vector.

\paragraph{Authored handle support.} Eq.~\ref{eq:motion-handle-set} defines the point set $\mathcal{H}_{t,i}$ using the fixed artist-selected indices $\mathcal{I}_i$. The selected points move as a rigid group under the handle's authored motion. Independently moving handles provide a piecewise-rigid approximation to non-rigid motion, and additional handles allow finer local control.

\section{Guidance Map Construction Details}
\label{supp:maps}

\paragraph{Shared coordinate normalization.} The background and foreground XYZ maps use $\mathcal{N}$ from Eq.~\ref{eq:xyz-normalization}, with shared parameters derived from the first-frame point cloud:
\begin{equation}
\begin{aligned}
\mathbf{c}_t&=\mathcal{N}(\mathbf{p}_t),\\
\mathbf{o}&=\frac{\mathbf{p}_{\max}+\mathbf{p}_{\min}}{2},
\qquad
\boldsymbol{\sigma}=\frac{\mathbf{p}_{\max}-\mathbf{p}_{\min}}{2},
\end{aligned}
\label{eq:supp:bg-norm}
\end{equation}
where $\mathbf{p}_t$ is a static world point for $\mathbf{F}^{\mathrm{bg}}$ or a current handle point for $\mathbf{F}^{\mathrm{fg}}$, and $\mathbf{c}_t$ is its encoded XYZ color as in Eq.~\ref{eq:xyz-values}. Before computing the first-frame extrema $\mathbf{p}_{\min}$ and $\mathbf{p}_{\max}$, we pull points beyond the $98$th depth percentile back to that depth along their viewing rays. This prevents distant outliers from inflating $\boldsymbol{\sigma}$. For an encoded point, the inverse affine map is $\mathbf{p}_t = (2\mathbf{c}_t-\mathbf{1}) \odot \boldsymbol{\sigma}+\mathbf{o}$, where $\odot$ denotes element-wise multiplication. We compute $(\mathbf{o},\boldsymbol{\sigma})$ once per clip and use it for both XYZ maps in every frame. Points within this bounding box map to $[0,1]^3$; positions outside it can exceed this range. Invalid pixels are set to $\mathbf{0}$. Clips with extreme $\boldsymbol{\sigma}$ are excluded from training.

\paragraph{Gaussian fitting for motion handles.} We approximate each transformed point set $\mathcal{H}_{t,i}$ by a 3D Gaussian $\mathcal{G}_{t,i}=(\boldsymbol{\mu}_{t,i},\boldsymbol{\Sigma}_{t,i})$ with empirical mean and covariance:
\begin{equation}
\boldsymbol{\mu}_{t,i} = \frac{1}{|\mathcal{H}_{t,i}|} \sum_{\mathbf{p} \in \mathcal{H}_{t,i}} \mathbf{p}, \qquad \boldsymbol{\Sigma}_{t,i} = \frac{1}{|\mathcal{H}_{t,i}|} \sum_{\mathbf{p} \in \mathcal{H}_{t,i}} (\mathbf{p} - \boldsymbol{\mu}_{t,i})(\mathbf{p} - \boldsymbol{\mu}_{t,i})^\top.
\label{eq:supp:gaussian}
\end{equation}
Handles with fewer than five points are filtered out of the training set. We regularize the covariance with a positive diagonal floor to stabilize the fit.

\begin{figure*}[t]
  \centering
  \includegraphics[width=\linewidth]{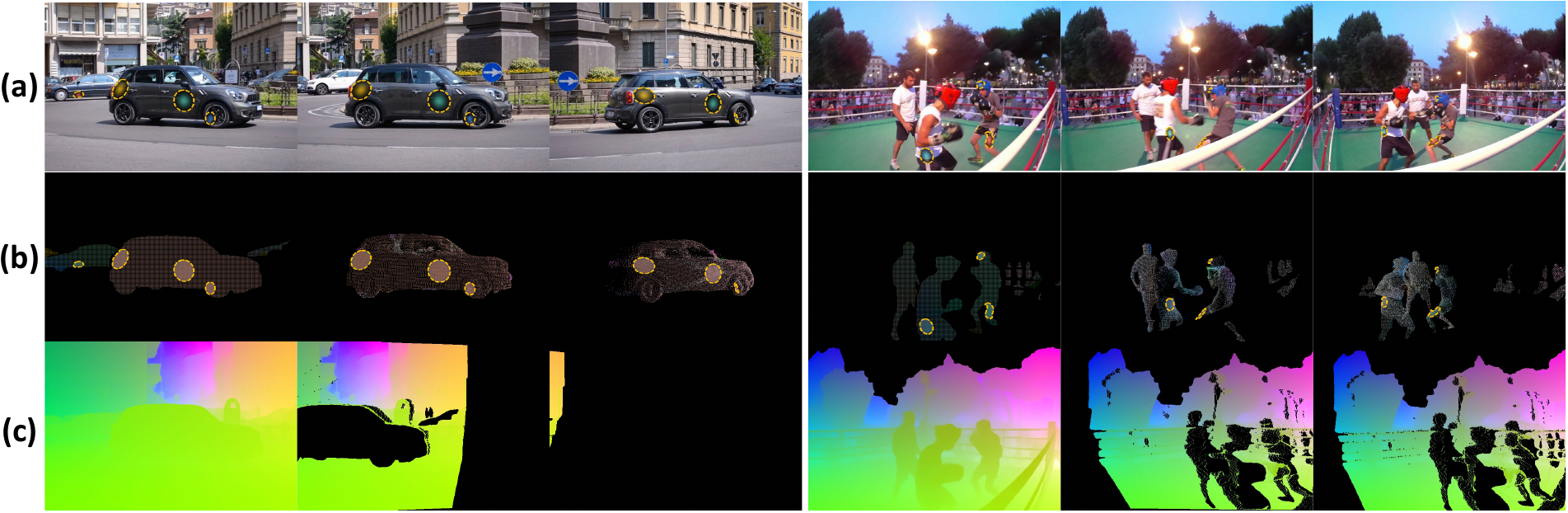}
  \caption{\textbf{Examples of the three guidance maps.}
  All maps use the same rasterized image-plane interface.
  \textbf{(a)} Foreground identity maps overlaid on RGB frames. Each artist-selected 3D handle region is rendered with a fixed identity color through its projected Gaussian footprint.
  \textbf{(b)} Foreground XYZ maps encode the current 3D positions of handle points as RGB values.
  \textbf{(c)} Background XYZ maps encode the world positions of valid static points. Both XYZ maps share one per-axis normalization, placing their values in the same coordinate system.}
  \label{fig:motion_maps}
\end{figure*}

\paragraph{Fixed identity colors.} When handle $i$ is created, we sample an identity color $\mathbf{c}_i\in[0,1]^3$ and keep it fixed throughout the video. Projecting $\mathcal{G}_{t,i}$ through Eq.~\ref{eq:projection} yields a 2D Gaussian with mean $\boldsymbol{\nu}_{t,i}$, covariance $\boldsymbol{\Lambda}_{t,i}$, and footprint $\Omega_{t,i}$. Within this footprint, the rendered value is $\alpha_{t,i}(\mathbf{u}_t)\mathbf{c}_i$ (Eq.~\ref{eq:fg-map}), where $\alpha_{t,i}\in[0,1]$ is largest at the center and decreases toward the boundary. Pixels outside all footprints are zero, and depth ordering selects the visible handle where footprints overlap. The foreground XYZ and identity maps share the same projected footprint, foreground validity mask, and visible handle. Fig.~\ref{fig:motion_maps} shows examples of all three guidance maps.

\paragraph{Validity masks.} Two pixel-wise validity masks accompany the three guidance maps. The background validity mask $\mathbf{M}^{\mathrm{bg}}$ marks pixels with valid rasterized background geometry, including supported holes filled by the refinement below. The foreground validity mask $\mathbf{M}^{\mathrm{fg}}$ marks pixels covered by a valid projected handle footprint and is shared by $\mathbf{F}^{\mathrm{fg}}$ and $\mathbf{F}^{\mathrm{id}}$. The first-frame dynamic mask $\mathbf{M}^{\mathrm{dyn}}$ provides a separate cue, with temporal padding as described in Sec.~\ref{supp:scaffold}. Invalid guidance-map pixels are zero; validity masks distinguish them from valid zero-valued encodings.

\paragraph{Occupancy refinement.} The background XYZ map is rasterized with a z-buffer and then morphologically refined. Closing only \emph{proposes} hole pixels; a proposed hole becomes valid only when a local 3D Gaussian fitted to the surrounding valid surface yields a camera-ray intersection. Unsupported holes remain zero and invalid. This local refinement does not reconstruct surfaces occluded in the input image.

\section{Architecture Details}
\label{supp:architecture}

\paragraph{Backbone.} We use Wan2.1-Fun-V1.1-14B-Control~\citep{wan2025wan} as the latent video diffusion backbone. The backbone consists of 40 DiT blocks operating on the video latent produced by Wan's VAE. The VAE and base diffusion weights are frozen; only the side branch, inter-branch projections, and main-branch LoRA adapters are trained.

\paragraph{Side branch.} The side branch follows the design of Wan-VACE~\citep{wan-vace}. It consists of 8 transformer blocks whose weights are initialized by copying from selected blocks of the main branch. The side branch contains approximately one-fifth of the main branch's parameter count.

The default side-branch spatial downsampling factor is $s_{\mathrm{side}}=2$: it operates at half the main branch's spatial resolution on the latent grid, using one quarter as many spatial tokens. This choice reduces conditioning cost and is motivated by the sparsity of the foreground XYZ and identity maps. Tab.~\ref{tab:quantitative_results2_2} compares it with $s_{\mathrm{side}}=1$.

\paragraph{Projection layers.} At the first side block, a Dirac-initialized strided $\mathrm{Conv3d}$ downsamples main-branch features and adds them to the encoded guidance features. Each side block emits features that a $\mathrm{ConvTranspose3d}$ projects back to the main branch's resolution. At an injection block $\ell$, the update is
\begin{equation}
\mathbf{x}_{\ell+1} = \mathrm{Block}_\ell(\mathbf{x}_\ell, \mathbf{y}) + s_{\mathrm{vace}} \cdot \mathbf{h}_\ell,
\label{eq:supp:residual}
\end{equation}
where $\mathbf{y}$ denotes Wan's text and diffusion-time conditioning, $\mathbf{h}_\ell$ is the projected side-branch feature, and $s_{\mathrm{vace}}$ is its scalar control strength. Output projections are zero-initialized, so the side branch initially contributes no residual. The main branch still receives the foreground identity sequence described below. Transposed-convolution output padding is determined from the requested main-branch grid to align feature dimensions across resolutions.

\paragraph{LoRA adaptation.} We additionally apply LoRA adapters~\citep{hu2022lora} to the main-branch attention and feed-forward layers ($q$, $k$, $v$, $o$, and both feed-forward projections), with rank $r = 64$. LoRA gives the main branch a small amount of capacity to adjust to the new conditioning distribution without disturbing the bulk of the pretrained weights.

\paragraph{Input channel layouts.} The main branch and side branch see different inputs.

\emph{Main branch.} Let $\tau$ denote diffusion time, distinct from video frame index $t$. The main branch concatenates the noisy video latent $\mathbf{z}_\tau$ with the encoded foreground identity sequence. The reference-frame latent $\mathbf{z}_{\mathrm{ref}}=\mathcal{E}(\mathbf{x}_0)$ enters as a fixed one-frame token prefix:
\begin{equation}
\text{main input} = \left[ \mathbf{z}_\tau \,\|\, \mathcal{E}(\mathbf{F}^{\mathrm{id}}) \,\|\, \mathbf{0} \right], \qquad \text{prefix} = \mathrm{RefConv}(\mathbf{z}_{\mathrm{ref}}).
\label{eq:supp:main-input}
\end{equation}
Here, $\mathbf{F}^{\mathrm{id}}$ denotes the full sequence of per-frame maps, and $\|$ denotes channel concatenation. The zero block pads the input to the Wan Control backbone's $48$ channels.

\emph{Side branch.} We reduce the binary masks to the VAE latent grid using max pooling, preserving small handle footprints during spatial and temporal reduction. Let $\widehat{\mathbf{M}}$ denote a processed mask tensor. The side branch concatenates the VAE-encoded guidance maps with these masks:
\begin{equation}
\begin{aligned}
\text{side input}=\bigl[&
\mathcal{E}(\mathbf{F}^{\mathrm{bg}})\,\|\,\widehat{\mathbf{M}}^{\mathrm{bg}}\,\|\,
\mathcal{E}(\mathbf{F}^{\mathrm{fg}})\,\|\,\widehat{\mathbf{M}}^{\mathrm{dyn}}\,\|\\
&\mathcal{E}(\mathbf{F}^{\mathrm{id}})\,\|\,\widehat{\mathbf{M}}^{\mathrm{fg}}\bigr].
\end{aligned}
\label{eq:supp:side-input}
\end{equation}
Each RGB guidance sequence contributes $C=16$ VAE channels, and each processed mask contributes $4$ channels in the input layout, for $16+4+16+4+16+4=60$ side-branch channels.

\paragraph{Text conditioning.} Text prompts are encoded by umT5~\citep{umt5} into text embeddings. The embeddings are injected into both the main branch and the side branch through Wan's standard text-conditioning interface; we make no modifications to the text pathway.

\section{Training Details}
\label{supp:training}

\paragraph{Data and adaptation.} As described in Sec.~\ref{sec:Experiment}, we train on $45$k clips, including approximately $5$k synthetic clips with ground-truth geometry and trajectories; real clips use estimated geometry and motion. Clips are processed at $480\times832$ resolution. We train the adaptation modules for approximately $20$k iterations on $4$ H100 GPUs.

\subsection{Sampling Motion Handles for Training}
\label{supp:handle-sampling}

For real clips, we estimate depth and camera trajectories with ViPE~\citep{huang2025vipe} and obtain motion trajectories with CoWTracker~\citep{lai2026cowtracker}, as in the main paper. Synthetic clips supply ground-truth geometry and trajectories. In both cases, we express foreground trajectories $\{\mathbf{p}_{t,j}\}_{t=0}^{T}$ in the canonical world frame and sample controls from the observed motion. Let $\mathcal{J}^{\mathrm{fg}}$ contain trajectories that have a finite, visible first-frame position projecting inside $\mathbf{M}^{\mathrm{dyn}}$. For each configuration, we sample the number of handles from a predefined range and select seed trajectories from $\mathcal{J}^{\mathrm{fg}}$.

Let $\mathbf{q}_{0,i}=\mathbf{p}_{0,j_i}$ be the initial center of motion handle $i$. We sample its radius relative to the scene scale $s$:
\begin{equation}
\begin{aligned}
\widehat{r}_i &\sim \mathcal{U}(r_{\min},r_{\max}), \\
r_i &= s\widehat{r}_i.
\end{aligned}
\label{eq:supp:handle-radius}
\end{equation}
The initial support contains all foreground trajectories within this radius:
\begin{equation}
\mathcal{I}_{0,i}
=\left\{j\in\mathcal{J}^{\mathrm{fg}}
\;\middle|\;
\lVert\mathbf{p}_{0,j}-\mathbf{q}_{0,i}\rVert_2\leq r_i
\right\}.
\label{eq:supp:initial-handle-support}
\end{equation}
We reject a new center if it lies inside the support ball of an accepted motion handle. This encourages different motion handles to cover distinct object parts.

Artist-authored handles use the fixed support $\mathcal{I}_i$ in Eq.~\ref{eq:motion-handle-set} and move rigidly. Training handles instead follow observed local tracks, whose support $\mathcal{I}_{t,i}$ may change as tracks become invalid or reappear. Thus, sampled training regions need not be exactly rigid. We update each handle independently. At frame $t\geq 1$, we collect valid trajectories near the previous center $\mathbf{q}_{t-1,i}$ into a candidate set $\mathcal{C}_{t,i}$. We fit an empirical mean $\boldsymbol{\mu}_{t,i}$ and a regularized covariance $\boldsymbol{\Sigma}_{t,i}$ to these candidates. For each $j\in\mathcal{C}_{t,i}$, the squared Mahalanobis distance is
\begin{equation}
d_{t,i}^{2}(j)
=
(\mathbf{p}_{t,j}-\boldsymbol{\mu}_{t,i})^{\top}
(\boldsymbol{\Sigma}_{t,i}+\epsilon\mathbf{I})^{-1}
(\mathbf{p}_{t,j}-\boldsymbol{\mu}_{t,i}).
\label{eq:supp:handle-mahalanobis}
\end{equation}
The refined support is $\mathcal{I}_{t,i}=\{j\in\mathcal{C}_{t,i}\mid d_{t,i}^{2}(j)\leq\tau_{\mathrm{M}}\}$. If the seed trajectory remains in this set, its position defines $\mathbf{q}_{t,i}$. Otherwise, when the refined support is nonempty, we use its mean position. We then refresh the candidate set around the updated center, allowing a temporarily missing trajectory to rejoin the motion handle in a later frame.

A frame is valid when $|\mathcal{I}_{t,i}|\geq n_{\min}$. Otherwise, it contributes a temporal gap. Let $\mathcal{V}_i=\{t\mid |\mathcal{I}_{t,i}|\geq n_{\min}\}$ denote the valid frames. We retain a motion handle only when its valid-frame ratio satisfies
\begin{equation}
\frac{|\mathcal{V}_i|}{T+1}\geq\rho_{\min}.
\label{eq:supp:handle-persistence}
\end{equation}
We also measure the mean normalized footprint area of the candidate handle,
\begin{equation}
A_i
=\frac{1}{|\mathcal{V}_i|}
\sum_{t\in\mathcal{V}_i}
\frac{|\Omega_{t,i}|}{HW},
\label{eq:supp:handle-area}
\end{equation}
and require $A_i\in[A_{\min},A_{\max}]$. This removes motion handles that are too small to be reliable or too large to represent local control. We resample highly overlapping motion handles until the desired number of distinct supports is obtained.

For every valid frame, we construct $\mathcal{H}_{t,i}=\{\mathbf{p}_{t,j}\mid j\in\mathcal{I}_{t,i}\}$ and rasterize it using the procedure in Sec.~\ref{supp:maps}. Invalid frames remain empty. We sample several handle configurations from each clip to vary the amount and location of foreground guidance. If $\mathcal{J}^{\mathrm{fg}}$ is empty, the foreground XYZ and identity maps and their shared validity mask are zero, leaving background XYZ guidance.

\paragraph{Training objective.} We adopt the flow-matching objective used by Wan:
\begin{equation}
\min_\phi \; \mathbb{E}_{(\mathbf{z}, g), \tau} \left[ \mathcal{L}_{\mathrm{FM}}(\mathbf{z}, \tau; c(g), \phi) \right],
\label{eq:supp:flow-matching}
\end{equation}
where $\mathbf{z}$ is the clean video latent from Eq.~\ref{eq:supp:latent}, $g$ is the precomputed geometry context, $c(g)$ is its conditioning tensor, and $\tau$ is diffusion time. Reference-frame and text conditioning are omitted from the notation for brevity. The parameters $\phi$ comprise the side branch, inter-branch projections, and main-branch LoRA adapters; all other parameters are frozen.

\section{Inference Details}
\label{supp:inference}

\paragraph{Artist interface.} The artist loads a single image into the Blender interface for \method. The system constructs the scene scaffold once from estimated depth, camera intrinsics, and the dynamic mask. In the 3D viewport, the artist selects foreground regions as motion handles and authors their trajectories together with a camera path using keyframes or drawn paths. Both controls use the same canonical world frame.

\paragraph{Partial guidance.} Sampling multiple handle configurations during training (Sec.~\ref{supp:handle-sampling}) exposes the model to different amounts of foreground guidance. With no foreground handles, $\mathbf{F}^{\mathrm{fg}}$, $\mathbf{F}^{\mathrm{id}}$, and $\mathbf{M}^{\mathrm{fg}}$ are zero. The model still receives the reference image, text, and background XYZ guidance under the chosen camera path; foreground motion is left to its learned prior. When no camera motion is authored, the camera remains at its initial pose.

\section{Failure Cases and Limitations}
\label{supp:failures}

\paragraph{Depth failure and occlusion.} Monocular depth estimation can fail on reflective or transparent surfaces and thin structures. Errors in the lifted point cloud propagate to the background XYZ map and can cause geometric inconsistencies during viewpoint changes. Even with accurate visible geometry, the single-image scaffold leaves occluded surfaces unmodeled; the generator must synthesize them as they become visible.

\paragraph{Piecewise-rigid approximation limits.} Each authored handle moves its selected points as a rigid group. Multiple handles can approximate articulated motion, but fracture, tearing, and fluid dynamics remain difficult. More handles allow finer control at the cost of additional authoring effort. Independently authored handles impose neither physical constraints nor causal interactions between objects, as discussed in Sec.~\ref{sec:discussion}.

\paragraph{Motion beyond the normalized range.} Both XYZ maps use the first-frame scene bounding box (Eq.~\ref{eq:supp:bg-norm}). Points within this box map to $[0,1]^3$, but a handle driven outside it can produce foreground XYZ values outside this range. Authoring within the scene extent keeps the guidance values in the RGB range; we expose the per-axis scale to the artist in the UI as a visual cue.

\end{document}